\RequirePackage{fix-cm}

\RequirePackage{snapshot}

\documentclass[twocolumn]{svjour3}          
\smartqed  
\usepackage{graphicx}
\usepackage{amsmath}
\usepackage{amssymb}
\newcommand{\minisection}[1]{\vspace{0.04in} \noindent {\bf #1}\ \ }
\usepackage{subcaption}
\usepackage{multirow}
\usepackage{multicol}

\usepackage{amsmath,amsfonts,bm}

\def\eqref#1{equation~\ref{#1}}

\def\1{\bm{1}}

\def\vw{{\bm{w}}}

\def\mS{{\bm{S}}}

\DeclareMathAlphabet{\mathsfit}{\encodingdefault}{\sfdefault}{m}{sl}
\SetMathAlphabet{\mathsfit}{bold}{\encodingdefault}{\sfdefault}{bx}{n}

\def\gF{{\mathcal{F}}}

\def\gW{{\mathcal{W}}}

\newcommand{\vdelta}   {\boldsymbol{\delta}}

\newcommand{\mH}{\mathbf{H}}
\newcommand{\vg}{\mathbf{g}}

\DeclareMathOperator*{\argmax}{arg\,max}

\usepackage[table]{xcolor}
\newcommand{\loss}{\mathcal{L}}

\usepackage[export]{adjustbox}%
\usepackage{tabularx}%
\usepackage{booktabs}  %
\usepackage{subcaption} %
\usepackage{array}%

\usepackage{array}
\makeatletter
\newcommand*{\@rowstyle}{}
\newcommand*{\rowstyle}[1]{
  \gdef\@rowstyle{#1}%
  \@rowstyle\ignorespaces%
}

\newcolumntype{=}{
  >{\gdef\@rowstyle{}}%
}

\newcolumntype{+}{
  >{\@rowstyle}%
}

\makeatother

\usepackage[round]{natbib}
\usepackage[final, colorlinks=true,allcolors=blue,breaklinks=true]{hyperref} 

\definecolor{f_trail}{RGB}{170,170,170}
\definecolor{f_grass}{RGB}{0,255,0}
\definecolor{f_vegetation}{RGB}{102,102,51}
\definecolor{f_sky}{RGB}{0,120,255}
\definecolor{f_obstacle}{RGB}{0,0,0}
\definecolor{u_Bed}{RGB}{0,0,255}
\definecolor{u_Books}{RGB}{233, 89, 48}
\definecolor{u_Ceiling}{RGB}{0, 218, 0}
\definecolor{u_Chair}{RGB}{149, 0, 240}
\definecolor{u_Floor}{RGB}{222, 241, 24}
\definecolor{u_Furniture}{RGB}{255, 206, 206}
\definecolor{u_object}{RGB}{0, 224, 229}
\definecolor{u_Picture}{RGB}{106, 136, 204}
\definecolor{u_Sofa}{RGB}{117, 29, 41}
\definecolor{u_Table}{RGB}{240, 35, 235}
\definecolor{u_Tv}{RGB}{0, 167, 156}
\definecolor{u_wall}{RGB}{250, 139, 0}
\definecolor{u_Window}{RGB}{225, 229, 195}

\begin{document}
\sloppy
\title{MineGAN++: Mining Generative Models for Efficient Knowledge Transfer to Limited Data Domains}

\author{Yaxing Wang,
        Abel Gonzalez-Garcia, Chenshen Wu,
        Luis Herranz, \\
        Fahad  Shahbaz Khan, Shangling Jui, Jian Yang and Joost van de Weijer
}


\institute{Y. Wang,  J. Yang   are VCIP, CS, Nankai University, Tianjin, China. \email{\{yaxing, csjyang\}@nankai.edu.cn}. \\
C. Wu, L. Herranz and J. van de Weijer are with the Computer Vision Center, Universitat Aut\`onoma de Barcelona, Barcelona 08193, Spain. \email{\{chenshen,lherranz,joost\}@cvc.uab.es}. \\
A. Gonzalez-Garcia is with WRNCH, Montreal,H4C 2Z6,  Canada. \email{abel.gonzalezgarcia@wrnch.ai}\\
F. Shahbaz  Khan  is   Mohamed bin Zayed University of Artificial Intelligence (UAE),  and Linkoping University (Sweden).  \email{fahad.khan@liu.se}\\
S. Jui  is  with Huawei Kirin Solution, Shanghai, China.\email{jui.shangling@huawei.com.}\\
}

\date{Received: date / Accepted: date}

\maketitle

\begin{abstract}
Given the often enormous effort required to train GANs, both computationally as well as in dataset collection, the re-use of pretrained GANs largely increases the potential impact of generative models. Therefore, we propose a novel knowledge transfer method for generative models based on mining the knowledge that is most beneficial to a specific target domain, either from a single or multiple pretrained GANs. This is done using a miner network that identifies which part of the generative distribution of each pretrained GAN outputs samples closest to the target domain. Mining effectively steers GAN sampling towards suitable regions of the latent space, which facilitates the posterior finetuning and avoids pathologies of other methods, such as mode collapse and lack of flexibility. Furthermore, to prevent overfitting on small target domains, we introduce \textit{sparse subnetwork selection}, that restricts the set of trainable neurons to those that are relevant for the target dataset. We perform comprehensive experiments on several challenging datasets using various GAN architectures (BigGAN, Progressive GAN, and StyleGAN) and show that the proposed method, called MineGAN, effectively transfers knowledge to domains with few target images, outperforming existing methods. In addition, MineGAN can successfully transfer knowledge from multiple pretrained GANs. \href{https://github.com/yaxingwang/MineGAN}{MineGAN}.

\end{abstract}

\section{Introduction}
\label{intro}
Generative adversarial networks (GANs) provide a powerful tool to learn complex distributions~\citep{goodfellow2014generative}. Given a target data distribution, a GAN aims to learn a generator network that can generate samples from that distribution. A discriminator, or critic, is learned simultaneously with the generator and aims to distinguish between real and generated target samples. The generator is optimized to trick the discriminator. They have been successfully applied in a wide range of applications, including image manipulation~\citep{pix2pix2017,zhu2017unpaired},  style transfer~\citep{gatys2016image}, compression~\citep{tschannen2018deep}, and colorization~\citep{zhang2016colorful}.

A known drawback of GANs is that they require large amounts of data to learn complex image data distributions~\citep{wang2018transferring,karras2020analyzing,brock2018large}. In addition, recent GAN architectures, that generate excellent high-quality realistic images, need excessive training times~\citep{karras2017progressive,karras2019style,karras2020analyzing,brock2018large}. For example, progressive GANs are trained on 30K images and require a month of training on one NVIDIA Tesla V100~\citep{karras2017progressive}.  Therefore, being able to exploit these high-quality pretrained models is desirable, not just to generate the distribution on which they are trained, but also to combine them with other models or adapt them to small target. For instance, one might only want to generate a subset of the original pretrained GAN, e.g. generating women using a pretrained GAN trained to generate men and women alike. Alternatively, one may want to generate smiling people from two pretrained generative models, one for men and one for women. In this paper, we study several different scenarios for transfer learning of GANs, and propose new techniques that allow for the re-use of pretrained generative models, thereby opening-up the usage of generative models to target domains with scarce data.

Transferring knowledge to domains with limited data has been extensively studied for discriminative models~\citep{donahue2014decaf,oquab2014learning,pan2010survey, tzeng2015simultaneous}, enabling the re-use of high-quality networks.
However, knowledge transfer for generative models has received significantly less attention, possibly due to its great difficulty, especially when transferring to target domains with few images.
Only recently, TransferGAN~\citep{wang2018transferring} studied finetuning from a single pretrained generative model and showed that it is beneficial for domains with scarce data. However, Noguchi and Harada~\citep{noguchi2019image} observed that this technique leads to mode collapse. 
Instead, they proposed to reduce the number of trainable parameters, and only finetune the learnable parameters for the batch normalization (scale and shift) of the generator. Despite being less prone to overfitting, their approach severely limits the effectiveness of the knowledge transfer. 

In this paper, we propose our method \emph{MineGAN} that addresses knowledge transfer by adapting a trained generative model for targeted image generation given a small sample of the target distribution.  
We introduce the process of \emph{mining} of GANs. This is performed by a \emph{miner network} that transforms a multivariate normal distribution into a distribution on the input space of the pretrained GAN in such a way that the generated images resemble those of the target domain. The miner network has considerably fewer parameters than the pretrained GAN and is therefore less prone to overfitting.  
The mining step predisposes the pretrained GAN to sample from a narrower region of the latent distribution that is closer to the target domain. This eases the subsequent finetuning step by providing a cleaner training signal with lower variance (in contrast to sampling from the whole source latent space as in TransferGAN~\citep{wang2018transferring}). 

In addition, our mining approach enables transferring from multiple pretrained GANs, which allows us to aggregate information from multiple sources simultaneously to generate samples akin to the target domain. We  study two approaches to exploit knowledge from multiple GANs. In the first approach, we show how to simultaneously mine multiple GANs; learning a separate miner network for each pretrained GAN. In the second approach, we consider first fusing the GANs into a single conditional GAN after which a standard mining operation on the single GAN can be performed. To the best of our knowledge, we are the first to consider transferring knowledge from multiple pretrained GANs to a target domain.

Furthermore, we present a new regularization of the GAN discriminator weights to further restrict overfitting. The proposed regularization is based on the idea that network weights have different levels of importance, and fixing the least important weights does not significantly reduce the capacity of the network to transfer to new domains.  Our regularization, called \emph{sparse subnetwork selection}, adaptively evaluates the importance of each filter based on its gradient norm, and then updates the important filters for the target domain and freezes the remaining filters.
Consequently, our method preserves the adaptation capabilities of finetuning while preventing overfitting. We call this improved version of our method \emph{MineGAN++}.

In summary, our main contributions are: 
\begin{itemize}
\setlength\itemsep{0em}
\item We introduce a novel miner network to steer the sampling of the latent distribution of a pretrained GAN to a target distribution determined by few images. 
\item We present a new regularization method, called sparse subnetwork selection, to localize  target-relevant parameters to reduce overfitting.
\item We are the first to address knowledge transfer from multiple GANs to a single generative model. 
\item Our approach performs favorably against existing methods on a variety of settings, including transferring knowledge from unconditional, conditional, and multiple GANs. 
\end{itemize}

This paper is an extension of our previous work~\citep{wang2020minegan}. We have included  more  analysis and  insight on  how to effectively leverage  pretrained generative models, as well as new techniques to improve their transfer, fusion and adaptation. In particular, we additionally include: (1) a new technique called \textit{sparse subnetwork selection} which formulates the importance of neurons of the pretrained model for target domain; (2)  a  \textit{fusion} mechanism that merges multiple pretrained GANs into a single model. This allows to utilize knowledge from all pretrained models;
(3) a new optimization procedure for knowledge transfer from multiple GANs that improves the performance of our preliminary work~\citep{wang2020minegan}. We also evaluate our method on the new pretrained StyleGAN~\citep{karras2019style}, which shows outstanding results for image generation. 
Furthermore, we conduct experiments and present results on two new datasets:  the \textit{Animal face}~\citep{si2011learning}, and the collected \textit{Winter church} dataset.

This paper is organized as follows. In next section, we review work from the literature. In Section~\ref{sec:uncondition} we explore the mining from
a single GAN, multiple GANs and elaborate on knowledge transfer with MineGAN.
In Section~\ref{sec:condition},  we introduce mining conditional GANs and fusion of mutiple pretrained GANs. We perform extensive  experiments to evaluate the proposed method in Section~\ref{sec:experiments}. Finally, conclusions are presented in Section~\ref{sec:conclusions}.

\section{Related work}

\subsection{Generative adversarial networks} GANs consists of two modules: a generator and discriminator, both of which play a minimax game~\citep{goodfellow2014generative}. 
The generator aims to synthesize a data distribution that mimics the real data distribution such that the discriminator fails to distinguish between images from the real distribution and generated images. 
GANs have been applied to various tasks: computer vision~\citep{dziugaite2015training,ma2017pose,vondrick2016generating}, and natural language processing~\citep{fedus2018maskgan,yang2017semi}, audio synthesization~\citep{donahue2018adversarial,liu2020unconditional}, etc. 

Training GANs, however,  is notoriously difficult, suffering from mode collapse (i.e., only generating a small subset of outputs) and training instability, since achieving a Nash equilibrium for both discriminator and generator is non-trivial. 
Currently, there are two main research directions in the GAN research community.  
The first trend focuses on improving the optimization method to more accurately match the real data distribution.
Some approaches~\citep{gulrajani2017improved,salimans2016improved,mao2017least,arjovsky2017wasserstein,miyato2018spectral} have successfully reduced training instability as well as mode collapse following this direction. Both WGAN~\citep{arjovsky2017wasserstein} and its variant the WGAN-GP~\citep{arjovsky2017iclr} leverage the Earth mover or Wasserstein-1~\citep{rubner2000earth} distance to optimize the training of GAN. LSGAN~\citep{mao2017least} introduces the decision boundary determined by the discriminator to address the vanishing gradient problem. Spectral normalization GAN (SN-GAN)~\citep{miyato2018spectral} applies the weight norm to optimize the training of the discriminator. RealnessGAN~\citep{xiangli2020real} proposes to use the multiple angles to estimate the \textit{realness} predicted by the discriminator. Sphere~GAN~\citep{park2019sphere}  presents a novel integral probability metric (IPM)-based GAN. This method defines IPMs on the hypersphere (i.e., a type of Riemannian manifolds), which can be trained stably using bounded IPMS.

The second direction attempts to find an effective architecture such that synthesized images are of the highest quality. 
Several architectures have been proposed to this end, with varying levels of success~\citep{radford2015unsupervised,denton2015deep,karras2017progressive,karras2019style,brock2018large}. 
The original GAN paper~\citep{goodfellow2014generative} employs a multiLayer perceptron (MLP) to train the generator and the discriminator.  
LAPGAN~\citep{denton2015deep} uses a multi-scale technique to generate higher resolution images. 
A major improvement was proposed by DCGAN~\citep{radford2015unsupervised}, which firstly applies the  transposed convolution for Generator. This method achieves excellent performance on image quality.  Recently, Progressive GAN~\citep{karras2017progressive} (PGAN) generates better images by synthesizing them progressively from low to high-resolution.  Two variants~\citep{karras2019style, karras2020analyzing} of PGAN  allow users to manipulate the output and reduce the number of artifacts in the generated images. 
MSG-GAN~\citep{karnewar2019msg} proposed a simple but effective technique to allow the flow of gradients from the discriminator to the generator at multiple scales. 
In terms of conditional image generation, BigGAN~\citep{brock2018large} successfully performs conditional high-realistic generation from ImageNet~\citep{deng2009imagenet}. Recent architecture-variant methods, however, present inferior performance when using limited data. In this paper, we aim to investigate how to synthesize high fidelity image from few images.

\subsection{Transfer learning for GANs}
While knowledge transfer has been widely studied for discriminative models in computer vision~\citep{donahue2014decaf,pan2010survey, oquab2014learning,tzeng2015simultaneous}. \citep{pmlr-v97-jitkrittum19a} explores the generative model, which provides a procedure for constructing a content-based generator by leveraging existing pretrained unconditional generative models. However, it does not focus on transfer learning for target data. 
TransferGAN~\citep{wang2018transferring} investigated finetuning of pretrained GANs, leading to improved performance for target domains with limited samples. This method, however, suffers from mode collapse and overfitting, as it updates all parameters of the generator to adapt to the target domain. Recently, Batch statistics adaptation (BSA)~\citep{noguchi2019image} propose to only update the batch normalization parameters. Although less susceptible to mode collapse, this approach significantly reduces the adaptation flexibility of the model since changing only the parameters of the batch normalization permits for style changes but is not expected to function when shape needs to be changed. They also replaced the GAN loss with a mean square error loss. As a result, their model only learns the relationship between latent vectors and sparse training samples, requiring the input noise distribution to be truncated during inference to generate realistic samples. Different to BSA~\citep{noguchi2019image},  concurrent work~\citep{zhao2020leveraging,mo2020freeze} only update a partial set of network parameters. Specifically, they fix the top layers (which are close to the output) of the pretrained generator, and the first layers of the pretrained discriminator. The two methods, however, require to select manually the specific layer which is frozen or not. Our method does not suffer from this drawback, as it learns how to automatically localize the target-specific weight. Another concurrent method~\citep{li2020few} proposes to combine pretrained GAN training with Elastic Weight Consolidation (EWC)~\citep{kirkpatrick2017overcoming}.
An additional difference between our work and existing work on transfer learning for GANs is that we are the first to consider transferring knowledge from multiple GANs to a single target domain.  

\subsection{GAN compression}
 {The effect of knowledge distillation on GANs has studied  the image-to-image translation. The authors of~\citep{shu2019co} propose the first end-to-end optimization framework combining multiple compression techniques for GAN compression. Naively minimizing the distance between the generated images of students and teachers leads to the  inferior performance of the student model~\citep{spkd_gan}. Li~\emph{et al.} introduce  the classification-based knowledge distillation for the image-to-image translation~\citep{gan_compress}. Recently, Chen~\emph{et al.}  propose an new knowledge distillation framework on GANs by distilling both the generator and the discriminator~\citep{distill_portable_gan}.
OMGD~\citep{ren2021online} introduce a GAN-oriented online scheme that the
teacher helps to warm up the student and guide the optimization direction step by step.  The work~\citep{Zhang_2022_CVPR} propose a  knowledge distillation method referred to as wavelet knowledge distillation to perform the knowledge distillation. Finally,  You~\emph{et al.}~\citep{you2022exploring} explores the content relationships for distillation of efficient GANs.}

\subsection{Transfer Learning in NLP}
  {In NLP, the transfer learning of pretrained language models has been widely explored.
Compared to fine-tuning, the parameter-efficient transfer method~\citep{houlsby2019parameter} obtains outstanding results with a small number of trainable parameters while achieving performance analogous to fine-tuning. Parameter-efficient transfer methods mainly consist of two directions: \textit{adapter-based} and \textit{prompt-engineering} approaches. The  adapter-based methods~\citep{houlsby2019parameter, pfeiffer2020adapterfusion} aim to search a small network, and freeze the pretrained language model.  The basic structure of the adapter-based method consists of a bottleneck layer, a nonlinear function, a normalization layer, and a residual connection. LoRA ~\citep{hu2021lora} inserts low-rank decomposition matrices for the weights in self-attention layers~\citep{vaswani2017attention}. To fully exploit the ability of the pretrained language model,   the prompt engineering approaches, transform the existing task into a text generation problem to predict the appropriate word in a given sentence. Previous works investigate handcrafted manual prompts ~\citep{schick2020exploiting, jiang2020can} or in-context learning~\citep{brown2020language, raffel2019exploring, gao2020making, zhao2021calibrate}. Since the drawback of the manual prompt, more recent works propose a soft prompt ~\citep{li2021prefix, lester2021power, shin2020autoprompt, liu2021gpt}. The soft prompt separates  the additionally trained parameters, and solely use the final output of the trained parameters as the prompt,  overcoming additional latency in the inference phase. In this paper, we  exploit similar ideas for parameter-efficient transfer learning for generative models.}

\subsection{Data augmentation } Data augmentation~\citep{shorten2019survey} is a data-space solution against overcome ovefitting when given limited data. For example, leveraging  various data processing techniques (e.g., flip, rotation, cutout, etc.,) results in increasing performance for classification task. Recently, data augmentation is also applied for GANs. Zhao et al.~\citep{zhao2020image,zhao2020improved} naively leverage  data augmentation to stabilize the training of GANs. DiffAugment~\citep{zhao2020diffaugment} and ADA ~\citep{karras2020training} show that differentiable augmentation techniques can  effectively stabilize training and vastly improve the quality when training data is sparse. They perform data augmentation for generated images as well as for the real one. As shown in ADA~\citep{karras2020training}, data augmentation obtains lower performance than transfer learning when the number of training  data is less than 1k (see Fig.9(c) in ~\citep{karras2020training}). In this paper, we focus on optimizing GANs on domains with fewer than 1k images.     

\subsection{Iterative image generation}
Nguyen et al.~\citep{nguyen2016synthesizing} have investigated training networks to generate images that maximize the activation of neurons in a pretrained classification network. 
In a follow-up approach~\citep{nguyen2017plug} that improves the diversity of the generated images, they use this technique to generate images of a particular class from a pretrained classifier network. 
In principle, these works do not aim at transferring knowledge to a new domain, and can instead only be applied to generate a distribution that is exactly described by one of the class labels of the pretrained classifier network. Another major difference is that the generation at inference time of each image is an iterative process of successive backpropagation updates until convergence, whereas our method is feedforward during inference.

\begin{figure*}[t]
    \centering
    \includegraphics[width=\textwidth]{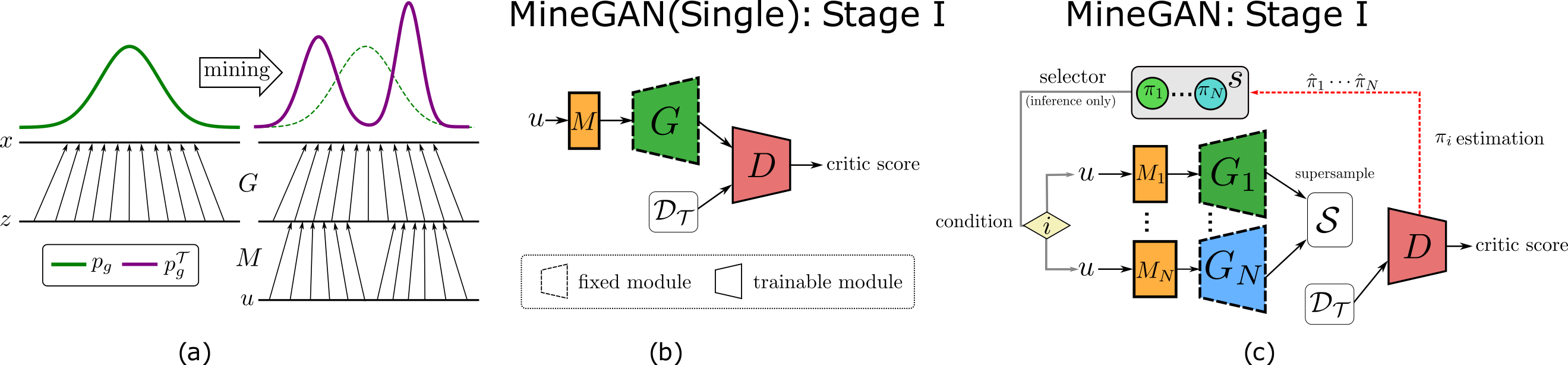}
    \caption{\small (a) Intuition behind our approach. Mining shifts the prior input distribution towards the most promising regions with respect to given target data $\mathcal{D}_\mathcal{T}$. In practice, the input distribution is much more complex. (b) Architecture implementing the proposed mining operation on a single GAN. Miner $M$ identifies the relevant regions of the prior distribution so that generated samples are close to the target data $\mathcal{D}_\mathcal{T}$. Note that during the first stage, when training the miner, the generator remains fixed; our method after only phase I is called MineGAN(w/o FT). In the second stage, we finetune the miner, generator and discriminator together. (c) Training setup for multiple generators. Miners $M_1$,...,$M_N$ identify subregions of the pretrained generators while selector $\mathcal{S}$ learns the sampling frequencies of the various generators. }
        \label{fig:models}
\end{figure*}

\section{Mining uncoditional GANs}\label{sec:uncondition}

Given one or more pretrained GANs and a few images of a target domain, our goal is to use the knowledge within the pretrained models to train a new GAN on the target domain in an effective fashion. We start with a general formulation of uncoditional GANs in Section~\ref{sec:formulationGAN}.
We then introduce mining from a single GAN in Section~\ref{sec:singleGAN}. Next, we present our general method for an arbitrary number of pretrained GANs in Section~\ref{sec:multipleGANs}. Finally, Section~\ref{sec:mineGAN} details how to apply the miners resulting from the previous sections to further improve the training of the new GAN models.

\subsection{GAN formulation}
\label{sec:formulationGAN}
Let $p_{data}(x)$ be a probability distribution over real data $x$ determined by a set of real images $\mathcal{D}$, and let $p_z(z)$ be a prior distribution over an input noise variable $z$.
The generator $G$ is trained to synthesize images given $z \sim p_z(z)$ as input, inducing a generative distribution $p_g(x)$ that should approximate the real data distribution $p_{data}(x)$.
This is achieved through an adversarial game~\citep{goodfellow2014generative}, in which a discriminator $D$ aims to distinguish between real images and images generated by $G$, while the generator tries to generate images that fool $D$.
More concretely, we follow the WGAN-GP~\citep{gulrajani2017improved} variant, which provides better convergence properties by using the Wasserstein loss~\citep{arjovsky2017wasserstein} and a gradient penalty term (omitted from our formulation for simplicity). The discriminator (also called critic) and generator losses are defined as follows:
\begin{equation}
\mathcal{L}_{D}= \mathbb{E}_{z \sim p_z(z)}[D(G(z))]
- \mathbb{E}_{x\sim p_{data}(x)}[D(x)] ,
\end{equation}
\begin{equation}
\mathcal{L}_{G} = - \mathbb{E}_{z \sim p_z(z)}[D(G(z))].
\end{equation}

In this paper, we also consider families of pretrained generators $\{G_i\}$.
Each $G_i$ has the ability to synthesize images given input noise $z^i\sim p_z(z)$.
For simplicity, and without loss of generality, we assume that the prior distributions are Gaussian, i.e.\ $p_z(z) = \mathcal{N}(z|\bm{\mu}, \bm{\Sigma})$.
Each generator $G_i(z)$ induces a learned generative distribution $p_g^i(x)$, which approximates the corresponding real data distribution $p_{data}^i(x)$ over real data $x$ given by the source domain image set $\mathcal{D}_i$.

\subsection{Mining from a single GAN} 
\label{sec:singleGAN}

We want to approximate a target real data distribution $p_{data}^\mathcal{T}(x)$ induced by a set of real images $\mathcal{D}_\mathcal{T}$, given a critic $D$ and a generator $G$, which have been trained to approximate a source data distribution $p_{data}(x)$ via the generative distribution $p_g(x)$.
The mining operation learns a new generative distribution $p_g^\mathcal{T}(x)$ by finding those regions in $p_g(x)$ that better approximate the target data distribution $p_{data}^\mathcal{T}(x)$ while keeping $G$ fixed.
In order to find such regions, mining actually finds a new prior distribution $p_z^\mathcal{T}(z)$ such that samples $G(z)$ with $z \sim p_z^\mathcal{T}(z)$ are similar to samples from $p_{data}^\mathcal{T}(x)$ (see Fig.~\ref{fig:models}a).
For this purpose, we propose a new GAN component called \emph{miner}, implemented by a small multilayer perceptron $M$. 
Its goal is to transform the original input standard Gaussian distribution noise variable $u \sim p_z(u)$ to follow a new, more suitable prior that identifies the regions in $p_g(x)$ that most closely align with the target distribution.

Our overall method comprises two stages.
The first stage steers the latent space of the fixed generator $G$ to suitable areas of the target distribution. 
We refer to the first stage as \emph{MineGAN (w/o FT)} and present the proposed mining architecture in Fig.~\ref{fig:models}b.
The second stage updates the weights of the generator via finetuning ($G$ is no longer fixed). 
\emph{MineGAN} refers to our overall method including finetuning.

\begin{figure}[t]
    \centering
    \includegraphics[width=1\columnwidth]{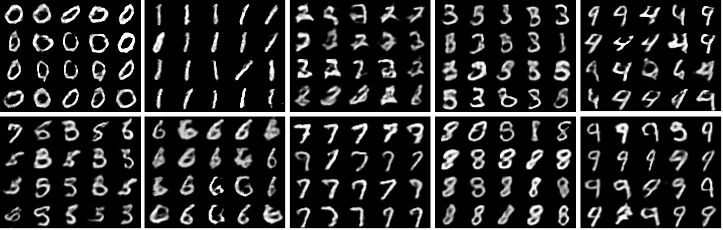}
    \caption{\small Results for far-domain generation of MineGAN (w/o FT). We generate 20 samples of digits $0\sim 9$. Taking digit '9'  as an example, the generated images are from the case where we train the miner using the pretrained model trained on digits '0'-'8', and take digit '9' as the target digit.   {Note that here} we only train the miner and do not finetune the generator on the target digit.}
    \label{fig:target_MNIST}
\end{figure}

Miner $M$ acts as an interface between the input noise variable and the generator, which remains trainable during training.
To generate an image, we first sample $u \sim p_z(u)$, transform it with $M$ and then input the transformed variable to the generator, i.e.\ $G(M(u))$. 
We train the model adversarially: the critic $D$ aims to distinguish between fake images output by the generator $G(M(u))$ and real images $x$ from the target data distribution $p^\mathcal{T}_{data}(x)$.
We implement this with the following modification on the WGAN-GP loss:
\begin{equation}
\mathcal{L}_D^{M}= \mathbb{E}_{u \sim p_z(u)}[D(G(M(u)))]
- \mathbb{E}_{x\sim p^\mathcal{T}_{data}(x)}[D(x)] ,\label{eq_min1}
\end{equation}
\begin{equation}
\mathcal{L}_{G}^{M} = - \mathbb{E}_{u \sim p_z(u)}[D(G(M(u)))].\label{eq_min2}
\end{equation}
The parameters of $G$ are kept unchanged but the gradients are backgropagated all the way to $M$ to learn its parameters.
This training strategy is expected to gear the miner towards the most promising regions of the input space, i.e.\ those that generate images close to $\mathcal{D}_\mathcal{T}$.
Therefore, $M$ is effectively mining the relevant input regions of prior $p_z(u)$ and giving rise to a targeted prior $p_z^\mathcal{T}(z)$, which will focus on these regions while ignoring other ones that lead to samples far off the target distribution $p_{data}^\mathcal{T}(x)$.

We distinguish two types of targeted generation: Close-domain and Far-domain.
In the \emph{Close-domain} case, there is a significant overlap between the original distribution $p_{data}(x)$ and the target distribution $p_{data}^\mathcal{T}(x)$.
For example, $p_{data}(x)$ could be the distribution of human faces (both male and female) while $p_{data}^\mathcal{T}(x)$ includes female faces only. 
On the other hand, in \emph{Far-domain} generation, the overlap between the two distributions is negligible, e.g.\ $p_{data}^\mathcal{T}(x)$ contains cat faces.
The Far-domain task is evidently more challenging as the miner needs to find samples out of the original distribution (see Fig.~\ref{fig:target_women_children}).
Specifically, we can consider the images in $\mathcal{D}$ to lie on a high-dimensional image manifold that contains the support of the real data distribution $p_{data}(x)$~\cite
{arjovsky2017iclr}.
For a target distribution farther away from $p_{data}(x)$, its support will be more disjoint from the original distribution's support, and thus its samples might be off the manifold that contains $\mathcal{D}$.
\begin{figure}[t]
    \centering
    \includegraphics[width=1\columnwidth]{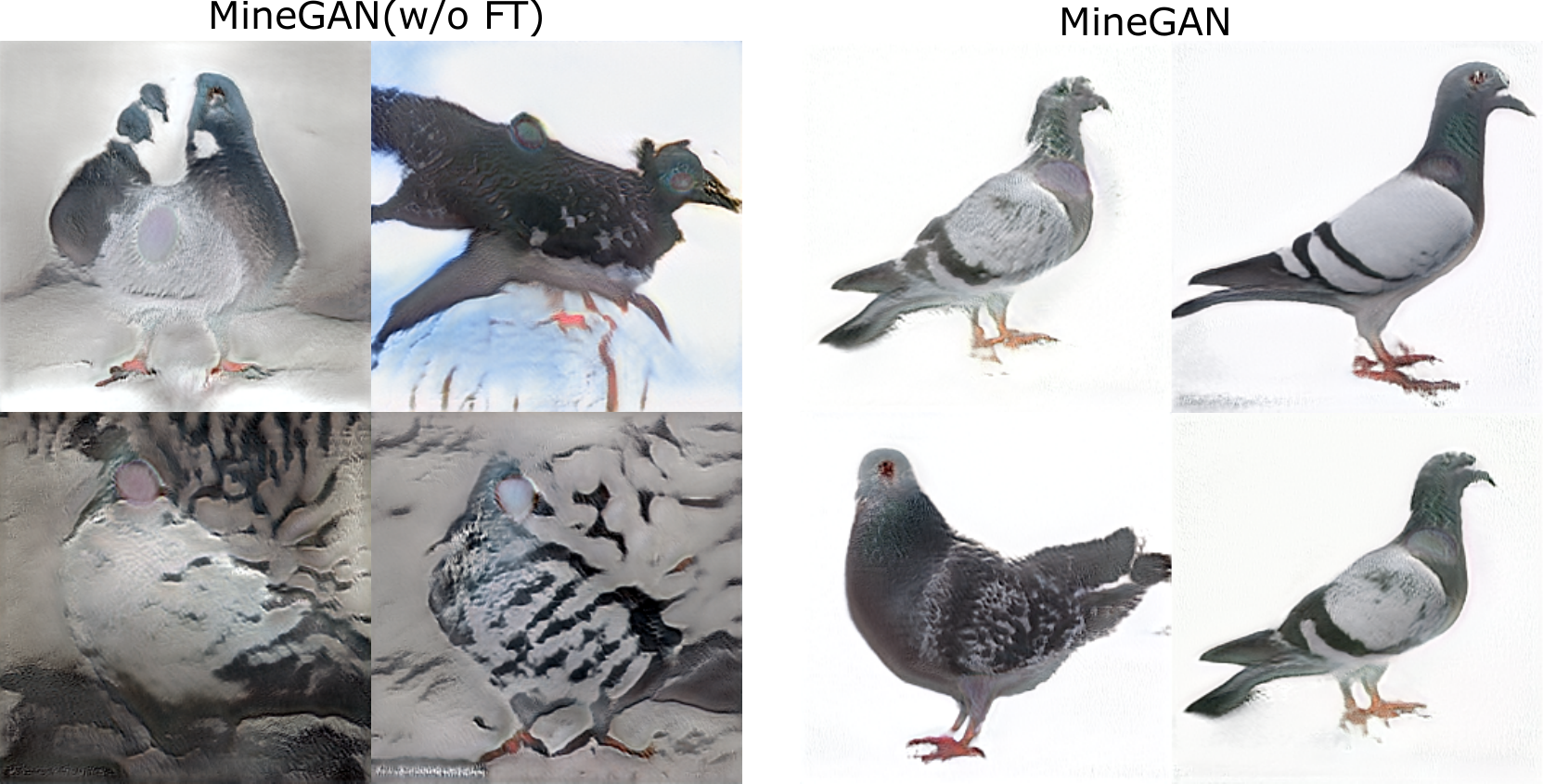}
    \caption{\small   { Illustration of  results of both MineGAN(w/o FT)(left) and MineGAN(right) transferring from a pretrained StyleGAN on human faces (FFHQ) to a target of 200 pigeons. Using the model from MineGAN(w/o FT) helps the subsequent finetuning.}}
    \label{fig:target_zebra}
\end{figure}

As an illustration of the mining process, we show results of an experiment on MNIST. We use 1000 images of size $28 \times 28$ as target data. We test mining for Far-domain targeted image generation.
In Far-domain targeted generation, $G$ is pretrained to synthesize all MNIST digits except for the target one, e.g.\ $G$ generates 0-8 but not 9.  The results in Fig.~\ref{fig:target_MNIST} are after training only the miner, without an additional finetuning step (MineGAN (w/o FT)).
Interestingly, the miner manages to steer the generator to output samples that resemble the target digits, by merging patterns from other digits in the source set.
For example, digit `9' frequently resembles a modified 4 while `8' heavily borrows from 0s and 3s. 
Some digits can be more challenging to generate, for example, `5' is generally more distinct from other digits and thus in more cases the resulting sample is confused with other digits such as `3'. In conclusion, even though target classes are not in the training set of the pretrained GAN, still similar examples might be found on the manifold of the generator.

  {As a second illustration we show results (Fig.~\ref{fig:target_zebra}) on a collected dataset containing 200 pigeon images. We use the unconditional pretrained  StyeGAN model optimized on FFHQ~\citep{karras2019style}. %
After the mining process (MineGAN (w/o FT)), we observe that the generated images are close to real pigeon, showing that our miner locates the related area in the latent space, which is closest to the target domain. We further  finetune the model, and obtain better generated images. } 

\begin{figure}[t]
    \centering
    \includegraphics[width=\columnwidth]{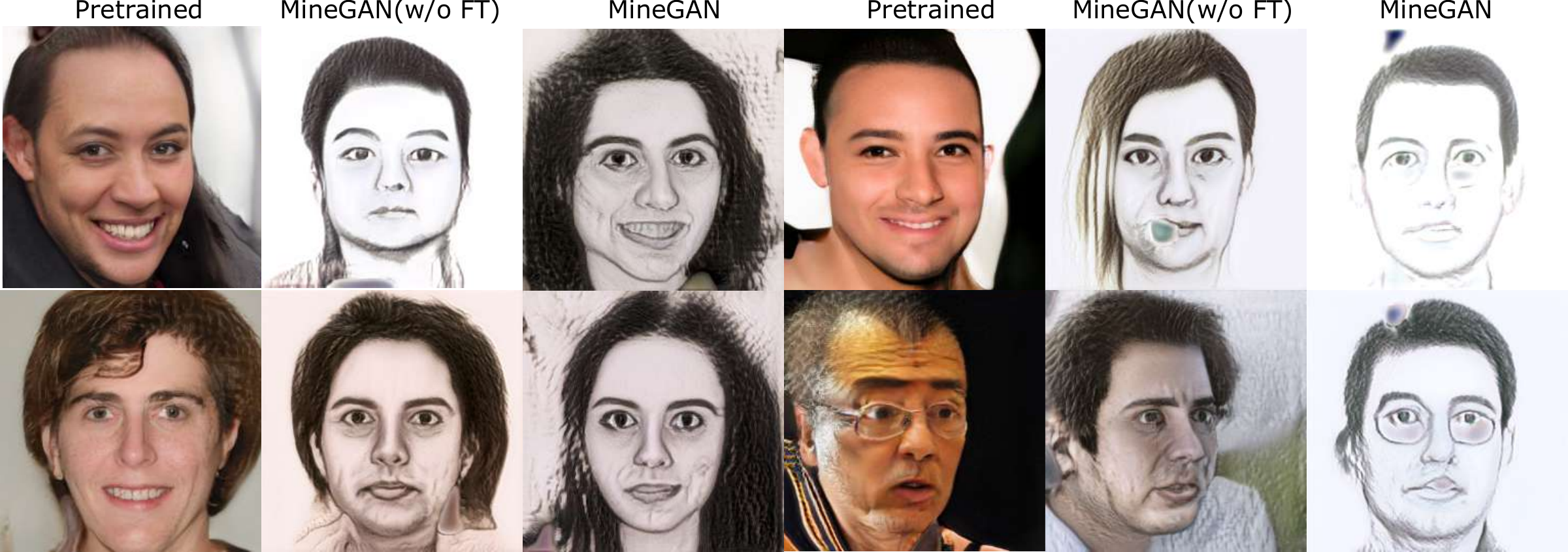}
    \caption{\small   {Illustration of the importance of finetuning when the domain shift to the target domain is large. Results of both MineGAN (w/o FT) and MineGAN on transferring from a pretrained StyleGAN on human faces (FFHQ) to a target of 500 portrait sketches. MineGAN(w/o FT)  helps the subsequent finetuning but does not yet obtains satisfying results.}} 
    \label{fig:sketches}
\end{figure}

As a third illustration we show results for MineGAN (w/o FT) when transferring from human face pretrained model to portrait sketches. we evaluate our method on this challenging setting: (source-FFHQ, target-sketch human face) using a  pretrained StyleGAN. We use 500 sketch faces with 256*256 resolution as a target domain. As shown in Figure~\ref{fig:sketches}, MineGAN (w/o FT) (which is only stage 1 of our method, without finetuning of the GAN on the target data) successfully retains the structure or geometry information of faces from the pretrained model, and discards the appearance information. The resulting images are not of high quality yet, but much closer to the  target sketch images. This is in accordance with our motivation: using the mining network to steer to the target-specific information. 

\subsection{Mining from multiple GANs} 
\label{sec:multipleGANs}
In the general case, the mining operation is applied on multiple pretrained generators.
Given target data $\mathcal{D}_\mathcal{T}$, the task consists in mining relevant regions from the induced generative distributions learned by a family of $N$ generators $\{G_i\}$.
In this task, we do not have access to the original data used to train $\{G_i\}$ and can only use target data $\mathcal{D}_\mathcal{T}$.
Fig.~\ref{fig:models}c presents the architecture of our model, which extends the mining architecture for a single pretrained GAN by including multiple miners and an additional component called \emph{selector}.
In the following, we present this component and describe the training process in detail.

\minisection{Supersample.}
In traditional GAN training, a fake minibatch is composed of fake images $G(z)$ generated with different samples $z\sim p_z(z)$.
To construct fake minibatches for training a set of miners, we introduce the concept of \emph{supersample}.
A supersample $\mathcal{S}$ is a set of samples composed of exactly one sample per generator of the family, i.e.\ $\mathcal{S} = \{G_i(z^i) | z^i\sim p_z(z); i=1,...,N\}$.
Each minibatch contains $K$ supersamples, which amounts to a total of $K\times N$ fake images per minibatch. 

\minisection{Selector.}
The selector's task is choosing which pretrained model to use for generating  samples during inference.
For instance, imagine that $\mathcal{D}_1$ is a set of `kitchen' images, $\mathcal{D}_2$ are `bedroom' images, and let  $\mathcal{D}_\mathcal{T}$ be `white kitchens'.
The selector should prioritize sampling from $G_1$, as the learned generative distribution $p_g^1(x)$ will contain kitchen images and thus will naturally be closer to $p_{data}^\mathcal{T}(x)$, the target distribution of white kitchens.
Should $\mathcal{D}_\mathcal{T}$ comprise both white kitchens and dark bedrooms, sampling should be proportional to the distribution in the data. 

We model the selector as a random variable $s$ following a categorical distribution parametrized by
$\pi_1,...,\pi_N$ with $\pi_i>0$, $\sum \pi_i = 1$.
We estimate the parameters of this distribution as follows.
The quality of each sample $G_i(z)$ is evaluated by a single critic $D$ based on its critic value $D(G_i(z))$. Higher critic values indicate that the generated sample from $G_i$ is closer to the real distribution. 
For each supersample $\mathcal{S}$ in the minibatch, we record which generator obtains the maximum critic value, i.e.\ $\argmax_{i} D(G_i(z))$.
By accumulating over all $K$ supersamples and normalizing, we obtain an empirical probability value $\hat{\pi}_i$ that reflects how often generator $G_i$ obtained the maximum critic value among all generators for the current minibatch. We estimate each parameter $\pi_i$ as the empirical average $\hat{\pi}_i$ in the last 1000 minibatches.
Note that $\pi_i$ are learned during training and fixed during inference, where we apply a multinomial sampling function  to sample the index.

\minisection{Critic and miner training.} We now define the training behavior of the remaining learnable components, namely the critic $D$ and miners $\{M_i\}$, when
minibatches are composed of supersamples. 
The critic aims to distinguish real images from fake images. This is done by looking for artifacts in the fake images which distinguish them from the real ones. 
Another less  discussed but equally important task of the critic is to observe the frequency of occurrence of images: if some (potentially high-quality) image occurs more often among fake images than real ones, the critic will lower its score, and thereby motivate the generator to lower the frequency of occurrence of this image.  
Training the critic by backpropagating from all images in the supersample prevents it from assessing the frequency of occurrence of the generated images and leading to unsatisfactory results empirically. 

To address this problem in~\citep{wang2020minegan} we proposed to use the following  loss for multiple GAN mining: 
\begin{equation}
\begin{split}
\mathcal{L}_D^{M}= \mathbb{E}_{ \{ u \sim p_z(u) \}}[\underset{i}{\max} \{D(G_i(M_i(u)))\}]\\
- \mathbb{E}_{x\sim p^\mathcal{T}_{data}(x)}[D(x)] 
\end{split}\label{eq:d_loss}
\end{equation}

\begin{equation}
\mathcal{L}_{G}^{M} = - \mathbb{E}_{\{u \sim p_z(u)\}}[\underset{i}{\max}\{D(G_i(M_i(u)))\}].\label{eq:g_loss}
\end{equation}
As a result of the $\max$ operator only the generator that resulted in the highest critic score is backpropagated. This allows the critic to assess the frequency of occurrence correctly. Using this strategy, the critic can perform both its tasks: boosting the quality of images as well as driving the miner to closely follow the distribution of the target set. 

However, this strategy suffers from two problems: 1. The generator is only updated when the output image obtains the higher scores, thus the low-score images that are not selected will not be improved. 2. At inference time, the selection of the GAN will be performed by the selector (and is not based on the max operator).
This asymmetry between training and inference reduces the quality of the final output.

Here we propose an improved method for the training of knowledge transfer to multiple GANs. We first rewrite Eqs.~\ref{eq:d_loss} and \ref{eq:g_loss}.
We sample $u(j) \sim p_u(u)$ where $j$ is the index of the supersample. Consider we define $m(j)=\underset{i}{\argmax} \{D(G_i(M_i(u(j)))\}$, then for the loss of a single minibatch Eqs.~\ref{eq:d_loss} and \ref{eq:g_loss} can be rewritten as:
\begin{equation}
\begin{split}
\mathcal{L}_D^{M}= \sum_{j=1}^K {D(G_{m(j)}(M_{m(j)}(u(j)))} \\
- \mathbb{E}_{x\sim p^\mathcal{T}_{data}(x)}[D(x)]
\end{split}\label{eq:d_loss_index}
\end{equation}

\begin{equation}
\begin{split}
\mathcal{L}_G^{M}= -\sum_{j=1}^K {D(G_{m(j)}(M_{m(j)}(u(j)))}
\end{split}\label{eq:g_loss_index}
\end{equation}

To solve the two above mentioned problems, we introduce a permutation operator $r(j)$ to permutate the index of supersample in the minibatch $1,2,...,K$ randomly after the selection of the miner. Like in the previous solution (Eqs.~\ref{eq:d_loss} and \ref{eq:g_loss}) we would like the critic to correctly assess the frequency of occurrence, but we want to remove the negative effects of propagating only from the images that obtain the maximum critic score. To do that, we apply the permutation operation.  This ensures that the number of times a generator is selected, and therefore will be updated by backpropagation, depends on how often it has generated the highest critic score in the supersample. This is important because it allows the critic to correctly assess whether the generated distribution   {will follow} the target distribution. But it will no longer only backpropagate the samples that generated these highest score, but because of the random nature of the permutation operation, it will also backpropagate some low-quality generated samples (improving the overall performance of the GANs). The losses are defined as:
\begin{equation}
\begin{split}
\mathcal{L}_D^{M} &= \sum_{j=1}^K {D(G_{m(r(j))}(M_{m(r(j))}(u(r(j))))} \\
&- \mathbb{E}_{x\sim p^\mathcal{T}_{data}(x)}[D(x)]
\end{split}\label{eq:d_loss_permutate}
\end{equation}

\begin{equation}
\begin{split}
\mathcal{L}_G^{M}= -\sum_{j=1}^K {D(G_{m(r(j))}(M_{m(r(j))}(u(r(j))))}
\end{split}\label{eq:g_loss_permutate}
\end{equation}
Now also the latent codes $u$ that do not directly result in the maximum critic score of the supersample can be updated through backpropagation. 

In the experimental section, we show that this newly proposed training scheme significantly improves over the one we proposed in~\citep{wang2020minegan}. Finally, we initialize the single critic $D$ in Eq.~\ref{eq:d_loss_permutate} with the pretrained weights from one of the pretrained critics. In the experiments, we find that the exact choice of pretrained critic is not that important (see ablation study in Section~\ref{sec:exp_mult_gans}).

\subsection{Knowledge transfer with MineGAN}
\label{sec:mineGAN}

Once we have found a suitable miner network that predisposes the GAN towards the target distribution, we can finetune the new GAN by releasing the weights of the generator. We discuss this beneficial process in Section~\ref{sec:finetuning}.
Nonetheless, further training the entirety of the model may potentially lead to undesirable overfitting, especially on small target datasets, as the number of trainable parameters is now substantially larger. 
For this reason, in Section~\ref{sec:S3} we propose a method to adaptively limit the number of trainable parameters in the network without compromising its performance, thereby mitigating the risk of overfitting to small target dataset.

\subsubsection{Finetuning MineGAN}\label{sec:finetuning}
The underlying idea of mining is to predispose the pretrained model to the target distribution by reducing the divergence between source and target distributions.
The miner network contains relatively few parameters and is therefore less prone to overfitting, which is known to occur when directly finetuning the generator $G$~\citep{wang2018transferring,noguchi2019image}. 
We finalize the knowledge transfer to the new domain by finetuning both the miner $M$ and generator $G$ (by releasing its weights). 
The risk of overfitting is now diminished as the generative distribution is closer to the target, requiring thus a lower degree of parameter adaptation.
Moreover, the training is substantially more efficient than directly finetuning the pretrained GAN~\citep{wang2018transferring}, where synthesized images are not necessarily similar to the target samples. A mined pretrained model makes the sampling more effective,  leading to less noisy gradients and a cleaner training signal.

\subsubsection{Sparse subnetwork selection}\label{sec:S3} \label{sec:filter_selection}
Although finetuning both the generator and the discriminator helps to reduce the divergence between source and target distribution, the discriminator may still suffer from overfitting.
In this case, training the large discriminator on the small target data produces very noisy gradients and leads the model to divergence~\citep{arjovsky2017iclr,zhang2019progressive}. To prevent such overfitting, we propose a method which reduces the number of trainable parameters. Our method, called \emph{Sparse Subnetwork Selection} ($S^3$), identifies a reduced number of trainable parameters in the pretrained discriminator. In particular, we select a sparse subset of filters that are highly sensitive to the target data. The remaining unselected filters are frozen to reduce the chance of overfitting. 

Our approach is inspired by the work of Frankle \& Carbin~\citep{frankle2018lottery} who propose the Lottery Ticket Hypothesis, suggesting that there exist sparse, trainable sub-networks (called ``winning tickets'')
within the larger network, that achieve comparable accuracy to the complete network. 
Later work~\citep{lee2019snip} found that these sub-networks can be automatically detected by considering the gradient norm flowing through the weights. Finally, this technique was improved by GraSP~\citep{wang2020picking} where 
they prune connections while taking their role in the network’s gradient flow into account.

In this paper, we show that the techniques developed in these papers can also be applied to select weights in pretrained networks to prevent overfitting. 
Whereas in~\citep{wang2020picking} they use their technique to identify neurons which can be removed from the network, we use a similar technique to identify the neurons which can be frozen during knowledge transfer to new domains. 
Specifically, we use an algorithm similar to GraSP to identify the target-specific filters based on the gradient norm. The main difference with GraSP is that, rather than selecting single weights, our method selects entire \emph{convolutional filters}.

Given the real data $x \sim p_{data}(x)$, the objective function of the critic $D$ parameterized by $\vw_0$ is formulated as $\loss^r_{D} = - \mathbb{E}_{x\sim p_{data}(x)}[D(x)]$. As in~\citep{wang2020picking}, we consider the gradient norm, which for a weight vector $\vw$ is given by:
\begin{equation}
\label{eq:loss_and_grad_norm}
\begin{aligned}
\Delta\loss^r_{D}(\vw) &= \lim_{\epsilon \rightarrow 0}\frac{\loss^r_{D}\left(\vw + \epsilon\nabla\loss^r_{D}(\vw)\right) - \loss^r_{D}(\vw)}{\epsilon} \\
&= \nabla\loss^r_{D}(\vw)^\top\nabla\loss^r_{D}(\vw).
\end{aligned}
\end{equation}

Here we not only consider the gradient in isolation, but follow~\citep{wang2020picking} and aim to preserve the flow of information through the network. Therefore, the norm of gradient is leveraged to preserve the reduction of the information flow. Specifically, we use a Taylor expansion to approximate how weights affect the gradient flow: 
\begin{equation}
\label{eq:s_func}
\begin{aligned}
    \mS\left(\vdelta\right) &= \Delta\loss^r_{D}(\vw_0 + \vdelta) - \Delta\loss^r_{D}(\vw_0)\\
    &= 2\vdelta^\top\nabla^2\loss^r_{D}(\vw_0)\nabla\loss^r_{D}(\vw_0)+ \mathcal{O}(\|\vdelta\|_2^2) \\
    &= 2\vdelta^\top \mH\vg + \mathcal{O}(\|\vdelta\|_2^2),
\end{aligned}
\end{equation}
where $\mH$ is the Hessian matrix, which reflects the dependencies between different weights.
For each weight $\vdelta$, the larger $S(\vdelta)$  the lower its importance.  Eq.~\ref{eq:s_func} can be further simplified to the following objective function: 
\begin{equation}
\centering
\label{eq:pruning_criteria}
    \begin{aligned}
         \mS(-\vw) = -\vw\odot \mH\vg.
    \end{aligned}
\end{equation}
We can now select the most relevant weights based on the values of $\mS(-\vw)$ on the target data. 
In order to determine which weights should be trained, we consider a threshold $\theta_w$ such that a desired number of parameters are selected as trainable.
Until here, we have followed~\citep{wang2020picking}, and thus we have a set of individual weights in the pretrained network to be trained.

In our approach, however, we aim to select entire filters (since the weights are tied together in convolutional filters). For this reason, we introduce a second threshold, $\theta_f$, indicating the minimum number of trainable weights a filter must contain to be considered trainable. 
Given a set of $K$ weights  $\gW = \{ w_1, w_2, ..., w_K \}$ in a filter $\gF$, we use the following criteria to select filters:
\begin{equation}
\label{eq:compar}
    \begin{aligned}
         \gF := \left\{\begin{matrix}
\text{1} & \text{if } \sum_{1}^{K}\left (\mS(w_{i})\ge \theta_{w}   \right )\ge \theta_{f} \\ 
\text{0} &  \text{otherwise}
\end{matrix}\right.
    \end{aligned}
\end{equation}
Where $0$ means that $\gF$ is frozen and $1$ trainable,  respectively. All weights which are not in the selected trainable filters are frozen. In the experiments (Fig.~\ref{fig:percentage_weight_on_layers} (c)), we show that our proposed method to select filters (based on Eq.~\ref{eq:compar}) obtains superior results to selecting weights (based on Eq.~\ref{eq:pruning_criteria}).  We found Sparse Subnetwork Selection to be beneficial in both stages of the MineGAN critic training: when training the miner and critic in the first stage and when training all weights jointly in the second stage.
We indicate that the MineGAN is finetuned using Sparse Subnetwork Selection with \emph{MineGAN++}.

\section{Mining conditional GANs}\label{sec:condition}
In the previous section, we have explored mining of unconditional pretrained GANs. However, many  GANs are conditional models, taking the class-label as an additional input. These models often synthesize higher quality images than the uncoditional GAN, showing that class-condition provides a valuable training signal~\citep{goodfellow2016nips}. Therefore, in Section~\ref{sec:mining_conditional}, we elaborate how to perform transfer learning with MineGAN on conditional GANs. To further leverage multiple pretrained GANs, we show how to fuse them into a single conditional GAN in Section~\ref{sec:fusion}. 
This is specially useful when combining GANs conditioned on categories that are not mutually exclusive, as it enables synthesising images conditioned on multiple categories.
For example, Fig.~\ref{fig:fusion_models} (b) illustrates the combination of pretrained GANs on \textit{Winter} and \textit{Church} categories into a new conditional GAN that can generate images of churches in winter.

\subsection{Mining Conditional GANs}\label{sec:mining_conditional} 
\begin{figure}[t]
    \centering
    \includegraphics[width=0.9\columnwidth]{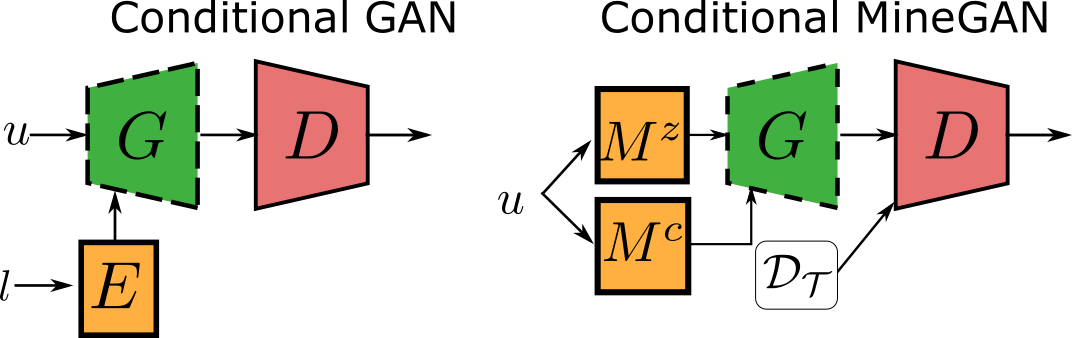}
    \caption{\small Application of mining in conditional setting (on BigGAN~\citep{brock2018large}). We apply an additional miner network to estimate the class embedding. $\mathcal{D}_\mathcal{T}$: target data, $E$: class embedding, $l$: label. }
    \label{fig:condition_models}
\end{figure}

Conditional GANs (cGANs), which introduce an additional input variable to condition the generation on the class label, are used by the most successful approaches~\citep{brock2018large,zhang2018self}.
Here we extend our proposed MineGAN to cGANs that condition on the batch normalization layer~\citep{dumoulin2016learned,brock2018large}, more concretely, BigGAN~\citep{brock2018large} (Fig.~\ref{fig:condition_models}~(left)).
First, a label $l$ is mapped to an embedding vector by means of a class embedding $E(l)$. Then this vector is mapped to layer-specific batch normalization parameters. The discriminator is further conditioned via label projection~\citep{miyato2018cgans}. 
Fig.~\ref{fig:condition_models}~(right) shows how to mine BigGANs.
Alongside the standard miner $M^z$, we introduce a second miner network $M^c$, which maps from $u$ to the embedding space, resulting in a generator $G(M^c(u),M^z(u))$. The training is equal to that of a single GAN and follows Eqs.~\ref{eq_min1} and~\ref{eq_min2}. Note we do not learn GMM~\citep{reynolds2009gaussian} over the input.  We introduce two miners for the conditional GAN, which has two inputs: the Gaussian noise and the class embedding.

\subsection{Fusing unconditional GANs into a conditional GAN}  
\label{sec:fusion}

In Sec.~\ref{sec:multipleGANs}, we show how to leverage the knowledge of multiple GANs. However, this method considers each generator in isolation, which fails to maximally exploit the knowledge of the multiple pretrained models. 
For example, when knowledge from multiple pretrained models is required to synthesize a target image, the previously described method is inadequate, since for the generation of a single image it can only extract knowledge from a single pretrained GAN.

\begin{figure}[t]
    \centering
    \includegraphics[width=0.9\columnwidth]{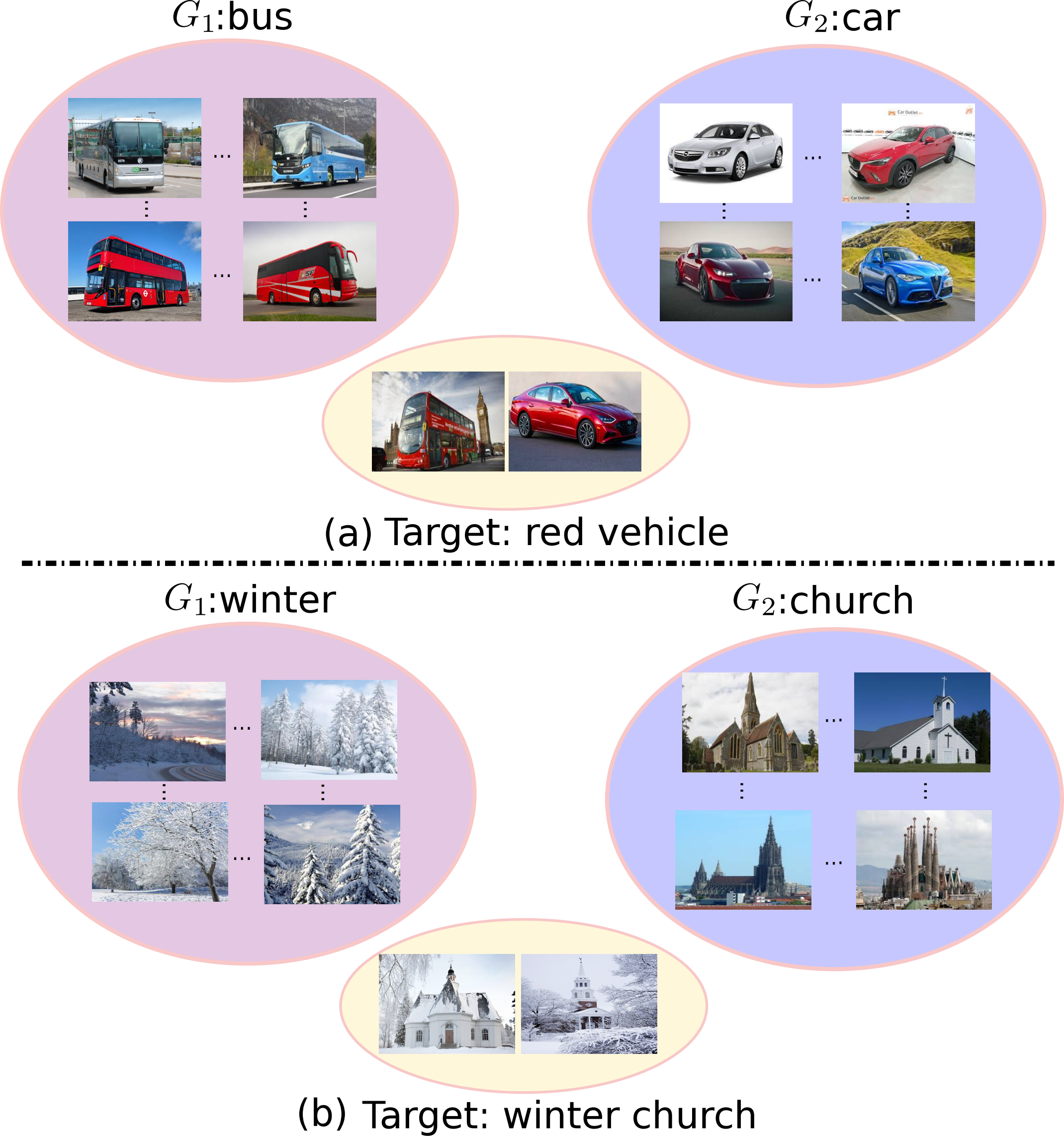}
    \caption{\small Application of multiple GANs for two cases of target data. (a) a single target image does not require simultaneously knowledge from the two pretrained GANs.  (b)~Knowledge from both pretrained GANs is required to synthesize effectively the target data. Zoom in for better details.}
    \label{fig:fusion_models}
\end{figure}

\begin{figure}[t]
    \centering
    \includegraphics[width=0.9\columnwidth]{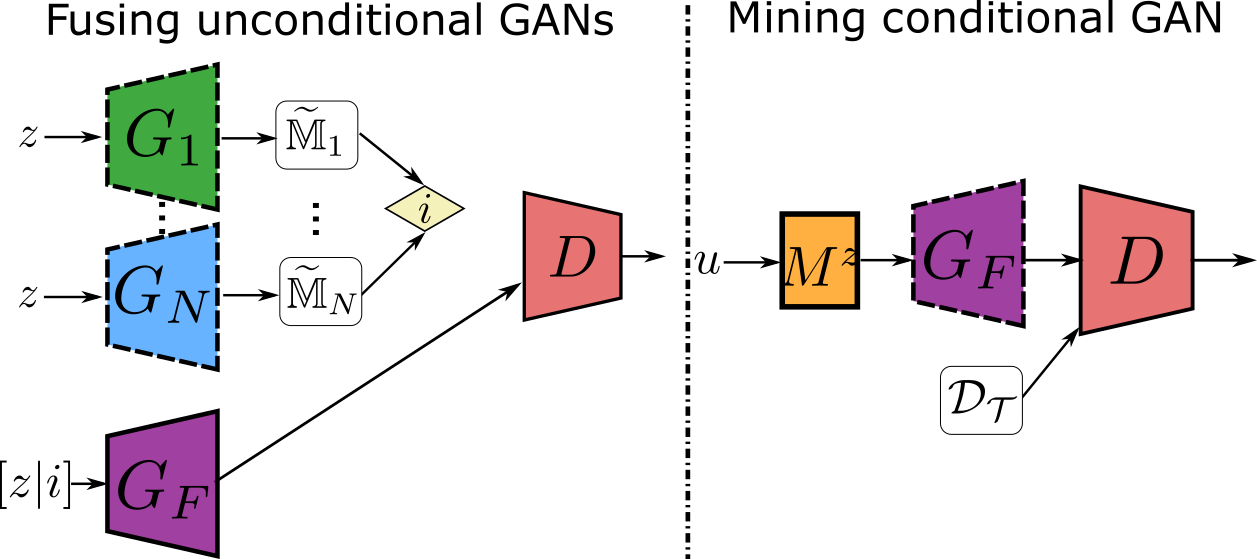}
    \caption{\small Framework of \textit{Fusion} and \textit{Mining}.  (Left) Fusion combines several pretrained models into a unique generator conditioned on the model. (Right) Miner $M^z$ identifies the relevant regions of the prior distribution (the fused generator $\mathbb{G}_F$), which is close to target data $\mathcal{D}_\mathcal{T}$.} 
    \label{fig:fusion_G}
\end{figure}

To further illustrate this point consider Fig.~\ref{fig:fusion_models}, which
 shows two cases of target data: Red vehicle and Winter church. As shown in  Fig.~\ref{fig:fusion_models}(a), each target image only has strong relation with one pretrained model (for example, the target red bus highly relies on bus ($G_{1}$) and only has a weak relation with car ($G_{2}$)).
However, each target sample of Fig.~\ref{fig:fusion_models}(b) simultaneously requires knowledge of both pretrained models, since the \textit{style} of a single target data depends on the Winter ($G_1$), while the \textit{content} has close correlation to Church ($G_2$). To allow the usage of the knowledge of multiple GANs for the generation of a single image, we propose to first fuse the multiple GANs into a single conditional GAN (Fig.~\ref{fig:fusion_G} (left)), and then execute the mining on the conditional GAN (Fig.~\ref{fig:fusion_G} (right)); this should allow the trained miner to exploit the information from both domains. It is important to note, that for the fusion, we will not require the training data of the pretrained models. 

We will now formalize fusing multiple unconditional GANs into a single conditional GAN.
Let us consider the set union of a collection of image sets, $\mathcal{D}_F = \cup_i \mathcal{D}_i$ and its induced fused manifold $\mathbb{M}_F$.
For example, if $\mathcal{D}_1$ is a set of `winter' images and $\mathcal{D}_2$ are `church' images, $\mathcal{D}_F$ is the set of images that is 'church and winter', which forms a new manifold $\mathbb{M}_F$ containing both $\mathbb{M}_1$ and $\mathbb{M}_2$.
The goal of the \emph{fusion} operation is to train a fused generator $G_F$ that approximates the fused manifold $\mathbb{M}_F$ without actually observing data $\mathcal{D}_F$, based solely on the approximated manifolds $\{\widetilde{\mathbb{M}}_i\}$ previously learned by generators $\{G_i\}$.

In order to implement the fusion operation, we propose using samples from the learned manifolds $\{\widetilde{\mathbb{M}}_i\}$ as \emph{pseudo real} data (Fig.~\ref{fig:fusion_G}(left)).
We train $G_F$ adversarially (e.g.~\citep{gulrajani2017improved}), using pseudo real data in the role normally taken by real data. The task of critic $D$ is assessing whether a particular sample has been generated by the fused model $G_F$ (i.e.\ the \emph{fake} data) or by one of the individual generators $G_i$ (\emph{pseudo real} data). 
More concretely, the  training objective now becomes
\begin{equation}
\label{eq:fusion_d_loss}
\begin{aligned}
\mathcal{L}_{D}^{F}&= \mathbb{E}_{z \sim p_z^F(z)}[D(G_F(z))]\\
&- \sum_i \mathbb{E}_{z\sim p_z^i(z)}[D(G_i(z))] \end{aligned}
\end{equation}
\begin{equation}
\label{eq:fusion_g_loss}
\mathcal{L}_{G}^{F} = - \mathbb{E}_{z \sim p_z^F(z)}[D(G_F(z))],
\end{equation}
where $p_z^F(z)$ be a prior distribution over the input noise variable $z$. We compute the expectations for each minibatch in a balanced manner, sampling from $p_z^F(z)$ as many times as the total number samples drawn from all $p_z^i(z)$ combined.
By optimizing Eq.~\ref{eq:fusion_d_loss} and~\ref{eq:fusion_g_loss}, the learned fused manifold $\mathbb{M}_F$ will contain all individual approximated manifolds, i.e.\ $\forall i$ $\widetilde{\mathbb{M}}_i \subseteq \mathbb{M}_F$.

In terms of probability distributions, we are training $G_F$ so that it maps input noise variable $z \sim p_z^F(z)$ to samples of the fused generative distribution $p_g^F(x)$.
Ideally, $p_g^F(x)$ should approximate the data distribution of $\mathcal{D}_F$.
Since we do not have direct access to $\mathcal{D}_F$, we are sampling from the generative distributions $\{p_g^i(x)\}$ via generators $\{G_i\}$.
With our proposed method, the learned $p_g^F(x)$ can be understood as a mixture distribution $\sum_i w_i p_g^i(x)$ with equal weights $w_i$ for all $i$ and $\sum_i w_i = 1$.
Fig.~\ref{fig:fusion_G}~(left) presents this intuition.
Based on this, $p_g^F(x)$ grants the fused generator $G_F$ the ability to output samples that lie on the individually learned manifolds $\{\widetilde{\mathbb{M}}_i\}$. 
Fig.~\ref{fig:fusion_G}~(right) shows that the fused generator is combined further with miner $M^z$, which is conducted as introduced in Section~\ref{sec:mineGAN}.    

In our paper, we consider the pretrained PGAN~\citep{zhang2019progressive}, but we remove the one-hot encoding because it results in inferior performance for high resolution images~\citep{mirza2014conditional}. 
Given that PGAN~\citep{karras2017progressive} does not apply batch normalization, conditioning via batch normalization as in Section~\ref{sec:mining_conditional} may result troublesome. 
Therefore, we apply here a different form of conditioning, using the index $i$ of the corresponding generator $G_i$ to condition $G_F$. Specifically, we leverage the Gaussian distribution with varying \textit{mean} and \textit{variance} corresponding  to different pretrained generators. Specifically, we assign a particular region of the input prior distribution to each pretrained generator. Each region is determined by an exclusive mean value of the Gaussian distribution. As such, conditioning happens implicitly in the input prior distribution by selecting the mean of the Gaussian from which to draw samples.

\begin{figure}[t]
    \centering
    \includegraphics[width=1\columnwidth]{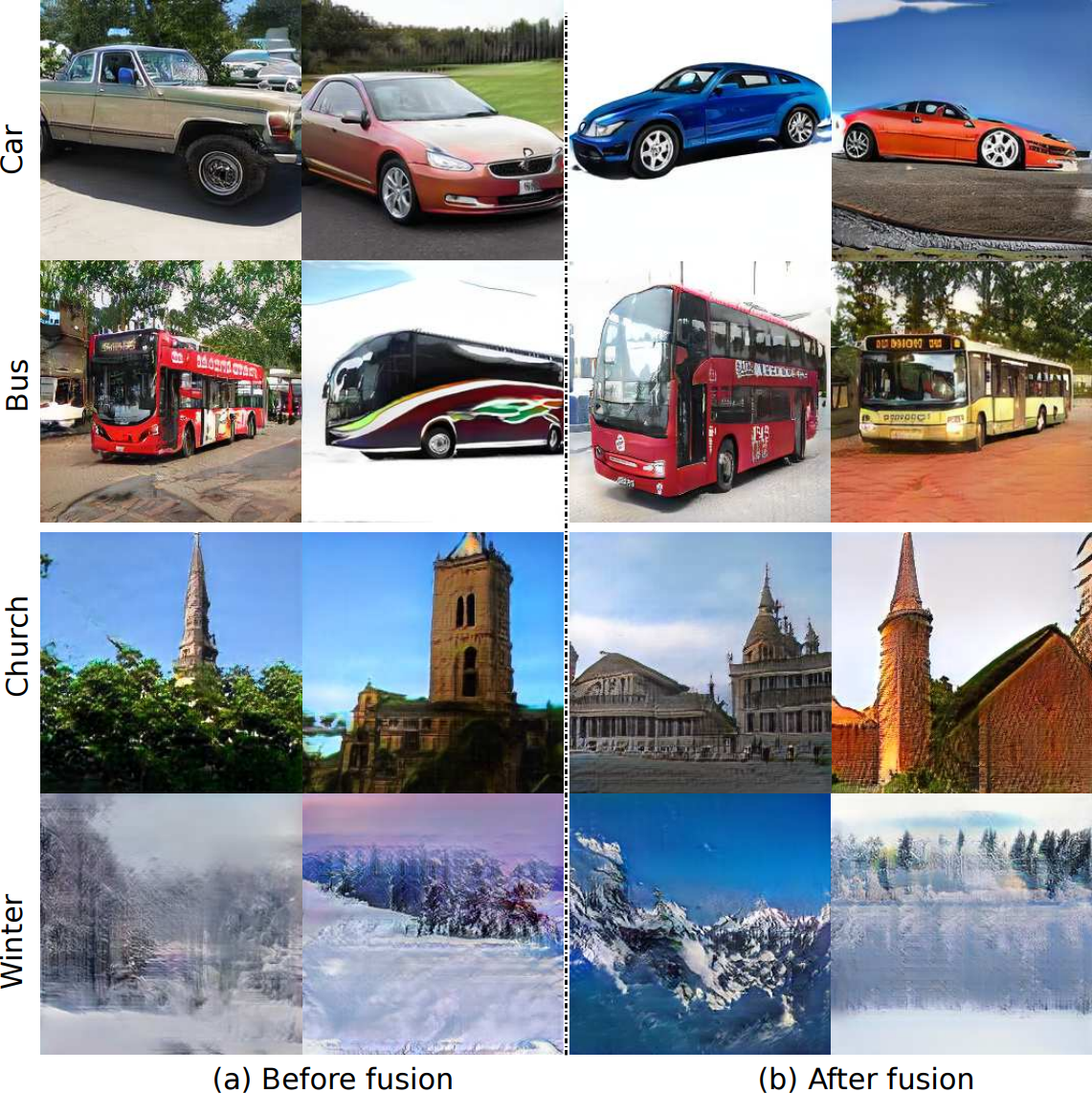}
    \caption{\small The qualitative result before and after fusion. Top rows show results for fusion of \textit{car} and \textit{bus} GANs into a single conditional GAN. Bottom rows show results of fusion of \textit{Church} and \textit{Winter}.}   \label{fig:car_bus_church_winter_fusion_comparsion}
\end{figure}

We  show samples of the generated images before and after fusion (see Fig.~\ref{fig:car_bus_church_winter_fusion_comparsion}). Here we show results for fusing two GAN models into a single conditional GAN. We fuse \textit{Car}, \textit{Bus} and 
 \textit{Church} and \textit{Winter} together. We can see that the images generated after fusion  have similar quality to those generated by the source models. We will perform mining on these fused models in the experimental section.

\begin{figure*}[t]
    \centering
    \includegraphics[width=1\textwidth]{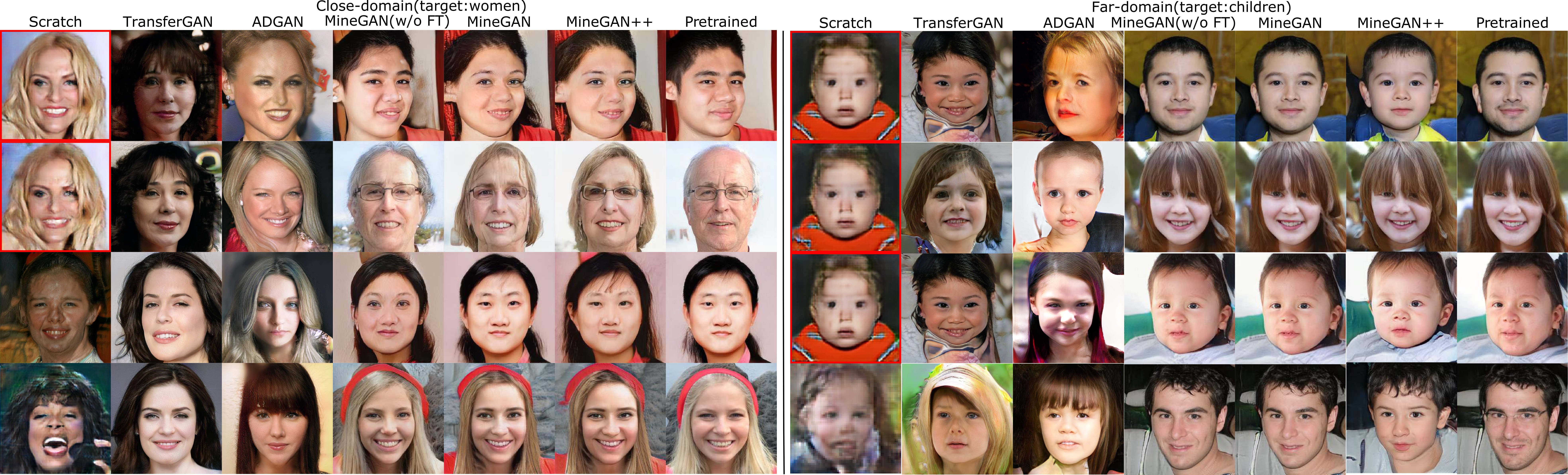}
    \caption{\small Results: (Left) Close-domain (CelebA-HQ $\rightarrow$FFHQ women),  (Right) Far-domain (CelebA-HQ $\rightarrow$FFHQ children). Based on pretrained Progressive GAN. The last column of each setting is from the pretrained PGAN. The images in red boxes suffer from overfitting.  }
    \label{fig:target_women_children}
\end{figure*}

\section{Experiments}
\label{sec:experiments}
We first introduce the evaluation measures and architectures used.
Then, we evaluate our method for knowledge transfer from unconditional GANs, considering both single and multiple pretrained generators. 
Finally, we assess transfer learning from conditional GANs. Experiments focus on transferring knowledge to target domains with few images. 

\minisection{Evaluation measures.} We employ the widely used Fr\'echet Inception Distance (FID)~\citep{heusel2017gans} for evaluation. FID measures the similarity between two sets in the embedding space given by the features of a convolutional neural network. 
More specifically, it computes the differences between the estimated means and covariances assuming a multivariate normal distribution on the features.
FID measures both the quality and diversity of the generated images and has been shown to correlate well with human perception~\citep{heusel2017gans}. 
Similarly to BSA~\citep{noguchi2019image}, we compute FID between 10,000 randomly generated images and 10,000 real images, if possible.
When the number of target image exceeds 10,000, we randomly select a subset containing only 10,000. 
On the other hand, if the target set contains fewer images, we cap the amount of randomly generated images to this number to compute the FID.

However, FID suffers from instability on small datasets. For this reason, we also employ Kernel Maximum Mean Discrepancy (KMMD) with a Gaussian kernel 
for some experiments~\citep{noguchi2019image}. In Appendix B we show that the ranking of methods with FID based on few images is consistent with the ranking based on larger test sets. This allows us to compare methods based on few images with the FID measure. It should be noted though that the absolute value of FID varies when varying the number of images used to compute it.

The Kernel Maximum Mean Discrepancy (KMMD) measure computes the dissimilarity between two probability distributions (the real data and fake data distribution) using samples drawn independently from each of them~\citep{fortet1953convergence}. The kernel MMD~\citep{gretton2012kernel} measures (square) MMD between the real and fake distribution for a fixed characteristic kernel function. Kernel MMD works surprisingly well when it operates in the feature space of a pre-trained CNN. It is able to distinguish generated images from real images, and both its sample complexity and computational complexity are low~\citep{gretton2012kernel,borji2019pros}. Low KMMD values indicate high quality images. When evaluating model with FID, we select the best snapshot according to the training set FID, and report FID evaluations with the test set for all methods that we compare, which also is used in related method~\citep{kumari2021ensembling}.

\minisection{Compared Methods.}
In our experiments, we compare a number of other methods for transfer learning of generative models:

\begin{itemize}
    \item {\bf MineGAN}: refers to our method and includes the two stages. Stage I which learns the miner while fixing the generator, and stage II which finetunes all networks on the target data.
    
    \item \textbf{MineGAN(w/o FT)}: refers to the results of our method only after stage I.
    
    \item MineGAN++: We use \emph{MineGAN++} to refer to MineGAN combined with the Sparse Subnetwork Selection (described in Sec.~\ref{sec:S3}). .
\end{itemize}

We refer with \emph{MineGAN} to our full method including finetuning, whereas \emph{MineGAN (w/o FT)} refers to only applying mining (fixed generator). We use \emph{MineGAN++} to refer to MineGAN combined with the Sparse Subnetwork Selection (described in Sec.~\ref{sec:S3}). In all experiments, we set  $\theta_w$ to the value which is corresponding to the top 30~\% of all weights, which is ranked using Eq.~\ref{eq:pruning_criteria}, and we set $\theta_f$ to 6. The parameters are fixed in the beginning and kept the same during training.

We compare our method with the following baselines.
\textit{TransferGAN~\citep{wang2018transferring}} 
directly updates both the generator and the discriminator for the target domain.
\textit{VAE~\citep{kingma2013auto}}
is a variational autoencoder trained following~\citep{noguchi2019image}, i.e.\ fully supervised by pairs of latent vectors and training images. 
\textit{BSA~\citep{noguchi2019image}} updates only the batch normalization parameters of the generator instead of all the parameters.
\textit{FreezeD~\citep{mo2020freeze}} performs simple finetuning of GANs with frozen lower layers of the discriminator. \textit{ADGAN~\citep{zhao2020leveraging}} not only freezes the lower layers of the discriminator, but also the upper ones of the generator. Furthermore, they propose additional layers to modulate the input layers of the generator and the output layers of the discriminator to boost the performance on small target datasets.
\textit{DGN-AM~\citep{nguyen2016synthesizing}} generates images that maximize the activation of neurons in a pretrained classification network.
\textit{PPGN~\citep{nguyen2017plug}} improves the diversity of DGN-AM  by of adding a prior to the latent code via denoising autoencoder.
Note that both of DGN-AM and PPGN require the target domain label, and thus we only include them in the conditional setting.

\minisection{Architectures.} We apply mining to  Progressive GAN~\citep{karras2017progressive}, SNGAN~\citep{miyato2018spectral},  StyleGAN~\citep{karras2019style} and BigGAN~\citep{brock2018large}. 
The miner has two fully connected layers for MNIST and four layers for all other experiments.
More training details in Appendix A.

\subsection{Knowledge transfer from unconditional GANs}
\label{sec:uncoditional}
We begin with the experiments under the unconditional setting, either from a single pretrained GAN (\ref{sec:exp_single_gan}) or multiple GANs (\ref{sec:exp_mult_gans}).

\subsubsection{Mining from a single GAN}
\label{sec:exp_single_gan}

\begin{table}[t]
\centering
	\def\arraystretch{0.5}
    \begin{adjustbox}{max width=\textwidth}
    
    \footnotesize{
    \begin{tabular}{m{0.5\textwidth}m{0.5\textwidth}}
   & \\         \resizebox{1\linewidth}{!}{
            \begin{tabular}{lcccc}
    \toprule
    \multirow{2}{*}{Method} & \multicolumn{2}{c}{\textit{FFHQ Women}} & \multicolumn{2}{c}{\textit{FFHQ Children}} \\
       & FID & KMMD    & FID & KMMD \\
     \midrule
    From scratch & 150.4 & 0.81& 168.6 & 0.96  \\
    TransferGAN~\citep{wang2018transferring} & 104.5 & 0.73 & 101.5 & 0.73 \\
    Freeze D~\citep{mo2020freeze}  & 101.6 & 0.72  & 97.4 & 0.70  \\
    ADGAN~\citep{zhao2020leveraging}  & 98.9 & 0.65  & 90.7 & 0.69  \\

    MineGAN (w/o FT) & 102.3 & 0.69  & 107.5 & 0.77  \\
    MineGAN & 93.1 & 0.58 & 93.4 & 0.68 \\ 
    MineGAN++ & \textbf{82.1} & \textbf{0.52} & \textbf{84.9} & \textbf{0.61} \\ 
                \bottomrule
            \end{tabular}
            }
    \end{tabular} }
    \end{adjustbox}
    \caption{\small Results for various knowledge transfer methods. Based on Progressive GAN. }
	\label{tab:metrics_FID_progress}
\end{table}

\begin{figure}[t]
    \centering
    \includegraphics[width=1\columnwidth]{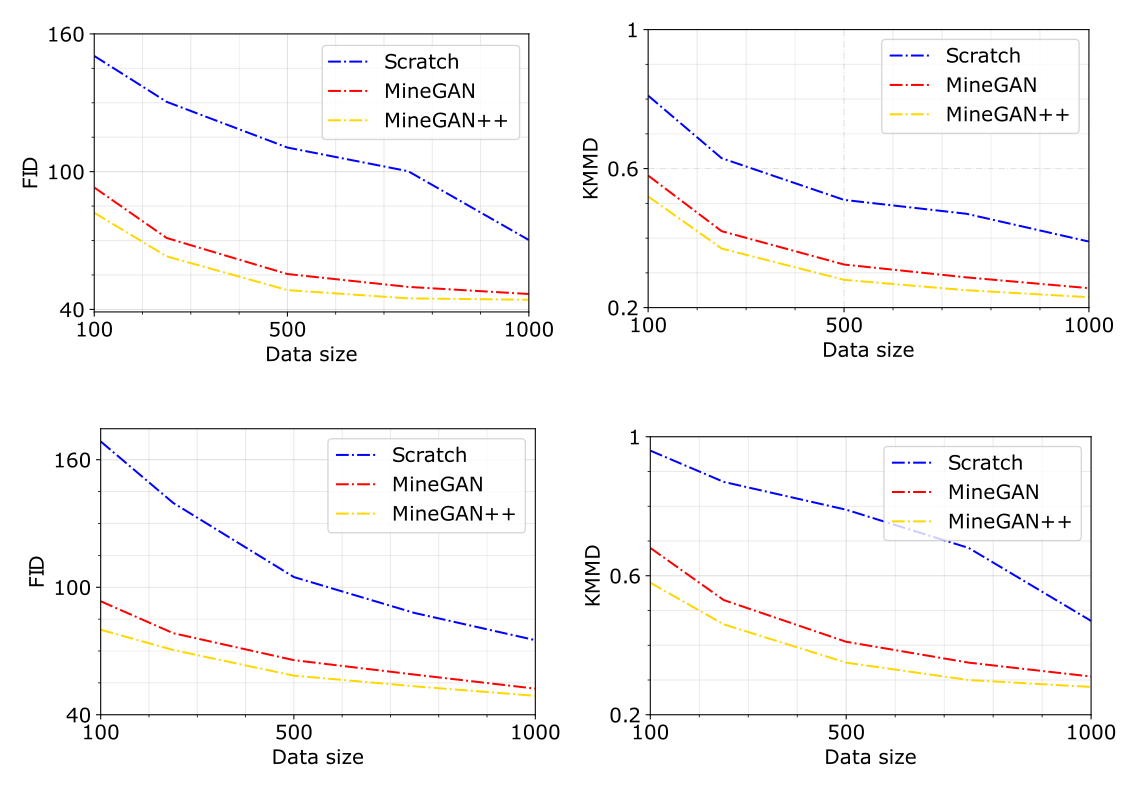}
    \caption{\small FID under different target size. (Top) FFHQ Children and (bottom) FFHQ Women. Note the pretrained model is trained on source domain (CelebA-HQ).}   \label{fig:target_women_children_fid}
\end{figure}

\minisection{Ablation study}

We ablate the influence of the number of the layers of M when combining with a pretrained Stylegan. As reported in Table ~\ref{tab:ablation_layer_miner}, given the specific training images the model performance gets worse when increasing the number of layers. The potential reason is that the increase of the layers requires more data to train, while, in this paper, we have limited data, which implies that less layers provide better results. Note that the mining network only combines fully connection layers (FC) and Relu. For example, 2 layers mean FC-Relu-FC with the corresponding size: (512, 512)-(512, 512).

Intuitively,  we introduce the mining network to explore and locate the related area of the latent space of the source domain with a small target domain. Given the limited target images, we prefer to design a small network. In conclusion, we use two small mining architectures throughout the experimental section, one when using the BigGAN and one when combined with the StyleGAN (details are provided in Appendix A).

\begin{table}[t]
\centering
	\def\arraystretch{0.5}
    \begin{adjustbox}{max width=\textwidth}
           \resizebox{1\linewidth}{!}{
            \begin{tabular}{lcc}
    \toprule
    \multirow{1}{*}{} & \multicolumn{1}{c}{100 training images} & \multicolumn{1}{c}{1000 training images} \\
     \midrule
    2 layers(512-512)
 & 51.6 & 46.0  \\
    4 layers(512-512-512-512) & 55.7 & 50.9 \\
    6 layers(512-512) & 58.3 & 52.4 
 \\ 
                \bottomrule
            \end{tabular}
            }
    \end{adjustbox}
    \caption{\small The ablation of the number of the layers  of M   {with metric FID}. We explore two cases: 100 and 1000 training cat images from AFHQ~\citep{choi2020stargan}. We use the pretrained StyleGAN to initialize our model, and evaluate our model with 500 test cat images of 256x256 resolution from AFHQ.} 
	\label{tab:ablation_layer_miner}
\end{table}

\minisection{FFHQ and Anime dataset.} We start by transferring knowledge from a Progressive GAN trained on \textit{CelebA-HQ~\citep{karras2017progressive}}. 
We evaluate the performance on target datasets of varying size, using $1024 \times 1024$ images. 
We consider two target domains: Close-domain, \textit{FFHQ women}~\citep{karras2019style} and Far-domain, \textit{FFHQ children face}~\citep{karras2019style}. 

We compare our method against training from \emph{Scratch}, \emph{TransferGAN}~\citep{wang2018transferring}, \textit{FreezenD}~\citep{mo2020freeze} and \textit{ADGAN}~\citep{zhao2020leveraging}. 
Table~\ref{tab:metrics_FID_progress} reports the compared result of the baseline and our method with 100 training images. Training a GAN from scratch obtains the worst performance on both Close-domain and Far-domain. ADGAN obtains better scores than MineGAN~(w/o FT) and all baseline methods. MineGAN achieves higher performance on \textit{FFHQ Women}, and obtains a competing score on \textit{FFHQ Children} when compared to ADGAN. Finally, MineGAN++  outperforms all baselines and MineGAN, clearly indicating that the proposed Sparse Subnetwork Selection is effective, reducing the FID score considerably by around 10 points on both settings.  Fig.~\ref{fig:target_women_children} shows images generated when the target data contains 100 training images. 
Training the model from scratch results in overfitting, a pathology also occasionally suffered by TransferGAN.
MineGAN++, in contrast, generates high-quality images without overfitting and images are sharper, more diverse, and have more realistic fine details.

Fig.~\ref{fig:target_women_children_fid} shows the performance in terms of FID and KMMD as a function of the number of target domain images. MineGAN significantly outperforms the model trained from scratch, and results are further improved with additional filter selection. 

\begin{figure}[t]
    \centering

    \includegraphics[width=0.5\textwidth]{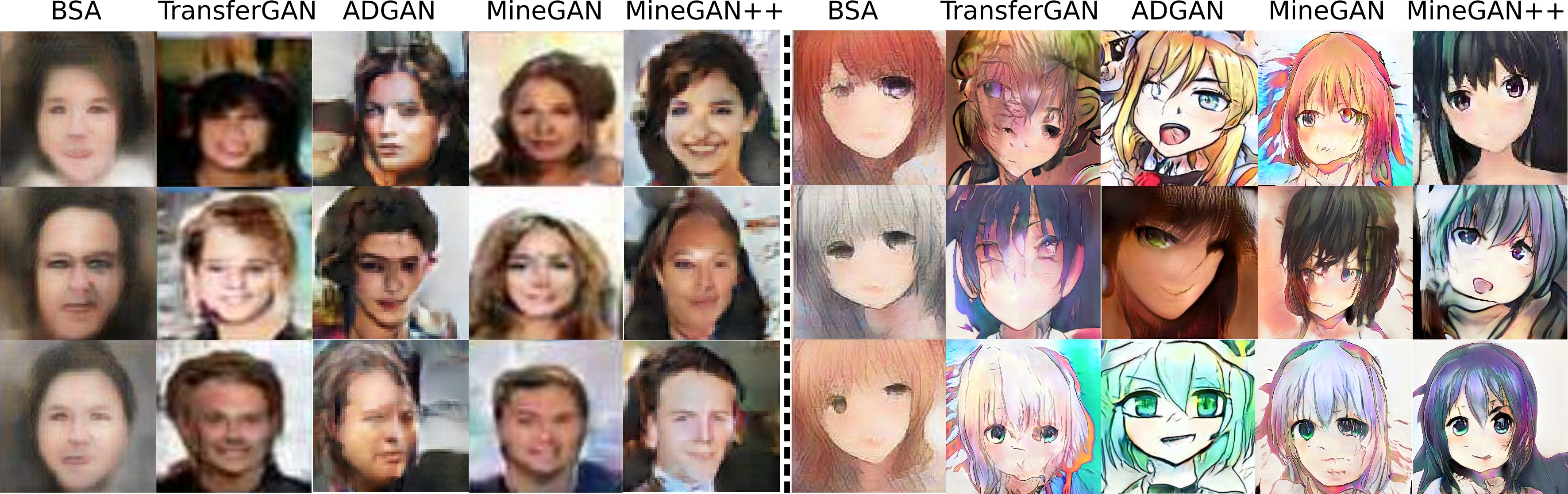}
    \caption{\small Qualitative result for various knowledge transfer methods on FFHQ (left) and Anime Face (right). }
    \label{fig:visualization_SNGAN}
    
\end{figure}

We also evaluate our method by using the pretrained model training by  SNGAN~\citep{miyato2018spectral}. Note that we use  the same settings and architecture used in  BSA~\citep{noguchi2019image}.
They performed knowledge transfer from a pretrained SNGAN on ImageNet~\citep{krizhevsky2012imagenet} to FFHQ~\citep{karras2019style} and to Anime Face~\citep{danbooru2018}. 
Target domains have only 25 images of size 128$\times$128. 
We added our results to those reported in~\citep{noguchi2019image} in Table~\ref{tab:metrics_KMMD_SNGAN}. 
Compared to BSA, MineGAN~(w/o FT) obtains similar KMMD scores, showing that generated images obtain comparable quality.   
MineGAN outperforms BSA both in KMMD score and Mean Variance. ADGAN obtains similar results like MineGAN. Finally, MineGAN++ achieves the best score on both metrics. The qualitative results (shown in Fig.~\ref{fig:visualization_SNGAN}) shows that MineGAN++ outperforms most baselines and MineGAN, and obtains similar results as ADGAN.
BSA presents blur artifacts, which are probably caused by the mean square error used to optimize their model.

\minisection{Animal dataset.} We further demonstrate the effectiveness of mining generative models on another target dataset. We leverage the pretrained StyleGAN model~\citep{karras2019style} trained on FFHQ~\citep{karras2019style} on Animal face~\citep{si2011learning}. Similar to FreezeD~\citep{mo2020freeze}, the 20 classes of the dataset are used with around 100 images per class after being resized to 256 $\times$ 256.  We take the publicly available model\footnote{https://github.com/rosinality/style-based-gan-pytorch.} of resolution 256 $\times$ 256 as our pretrained model.

\begin{table}[t]
\centering
	 \setlength{\tabcolsep}{15pt}\renewcommand{\arraystretch}{1.2}

    \begin{adjustbox}{max width=\textwidth}
           \resizebox{1\linewidth}{!}{
            \begin{tabular}{lcc}
    \toprule
    \multirow{2}{*}{Method} & \multicolumn{1}{c}{FFHQ} & \multicolumn{1}{c}{Anime Face} \\
       & KMMD     & KMMD  \\
     \midrule
    From scratch & 0.890 &  0.753   \\
    TransferGAN~\citep{wang2018transferring} & 0.346  & 0.347  \\
    VAE~\citep{kingma2013auto} & 0.744 &  0.790  \\
    BSA~\citep{noguchi2019image} & 0.345  & 0.342    \\
    ADGAN~\citep{zhao2020leveraging} & 0.341  & 0.330    \\\midrule
    MineGAN (w/o FT) & 0.349   & 0.347   \\
    MineGAN & 0.337  & 0.334 \\ 
    MineGAN++ & \textbf{0.329}  & \textbf{0.322}   \\ 
                \bottomrule
            \end{tabular}
            }
    \end{adjustbox}
    \caption{\small Results for various knowledge transfer methods. Based on SNGAN.}
	\label{tab:metrics_KMMD_SNGAN}
\end{table}

\begin{table*}
\resizebox{\textwidth}{!}{%
\begin{tabular}{lcccccccccc}
	\toprule
	            &Bear & Cat & Chicken & Cow & Deer & Dog & Duck & Eagle & Elephant & Human\\
	\midrule
	Scratch & 330.1 & 321.5 & 350.3 & 315.8 & 345.6 & 346.7 & 352.7 & 332.7 & 329.5 & 384.6\\
	TransferGAN  & 82.82 & 71.76 & 88.10 & 87.07 & 82.11 & 64.28 & 92.54 & 85.52 & 84.10 & 76.62\\
	TransferGAN++ & 77.84 & 69.05 & 82.54 & 83.40 & \textbf{77.23} & 61.49 & 86.94 & 81.52 & 79.84 & 71.86\\
	FreezeD & 78.77 & 69.64 & 86.20 & 84.32 & 78.67 & \textbf{61.46} & 88.82 & 82.15 & 80.00 & 73.51\\
	ADGAN & 78.50 & 69.57 & 83.48 & 82.04 & 147.2 & 62.57 & 88.86 & 82.51 & 80.87 & 72.04\\
	MineGAN & 76.53 & \textbf{68.10} & 83.28 & 82.36 & 150.4 & 86.41 & 88.72 & 79.31 & 78.83 & 71.03\\
	MineGAN++ & \textbf{75.07} & 71.09 & \textbf{80.98} & \textbf{79.84} & 139.1 & 84.87 & \textbf{86.54} & \textbf{75.90} & \textbf{76.07} & \textbf{68.43}\\
	\midrule
	            & Lion & Monkey & Mouse & Panda & Pigeon & Pig & Rabbit & Sheep & Tiger & Wolf\\
	\midrule
	Scratch  & 320.4 & 325.6 & 341.2 & 310.9 & 328.4 & 336.4 & 300.4 & 384.5 & 315.4 & 327.6\\
	TransferGAN  & 76.86 & 86.70 & 84.95 & 74.29 & 81.24 & 85.31 & 89.11 & 86.98 & 73.21 & 79.97\\
	TransferGAN++ & 72.64 & 82.26 & 79.41 & 71.73 & 75.89 & 83.41 & 83.68 & 82.59 & 70.29 & 74.84\\
	FreezeD & 73.49 &  82.31 &  81.72 &  72.19 &  77.79 &  83.22 &  85.65 &  84.33 &  71.26 &  76.47\\ 
	ADGAN & 72.58 &  82.06 &  80.54 &  71.48 &  75.04 &  75.97 &  83.88 &  82.04 &  70.59 &  74.93\\ 
	MineGAN & 72.39 & 82.10 & 80.12 & 70.46 & 75.39 & 80.00 & 83.35 & 83.05 & 71.66 & \textbf{74.55}\\
	MineGAN++ & \textbf{70.27} & \textbf{79.53} & \textbf{78.54} & \textbf{69.31} & \textbf{74.63} & \textbf{78.60} & \textbf{81.43} & \textbf{79.56} & \textbf{69.20} & 74.87\\
	\bottomrule
\end{tabular}}
\caption{FID scores under Animal face dataset. Our MineGAN obtains best results on 18 out of 20 classes. } \label{tab:fid_animal}
\end{table*}

\begin{figure*}[t]
    \centering
    \includegraphics[width=\textwidth]{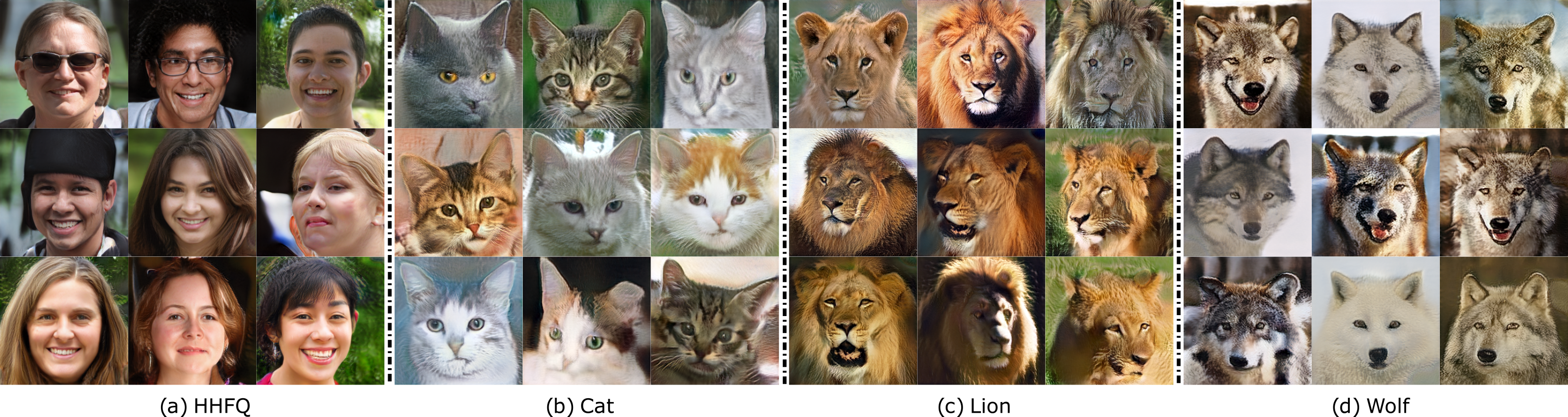}
    \caption{\small   {MineGAN++} results generated by StyleGAN. (a) the randomly generated sample from the pretrained StyleGAN on FFHQ dataset. The synthesized sample on target domains: \textit{cat} (b), \textit{lion} (c) and \textit{wolf} (d). }
    \label{fig:stylegan_animal}
\end{figure*}

\begin{figure*}[t]
    \centering
    \includegraphics[width=1\textwidth]{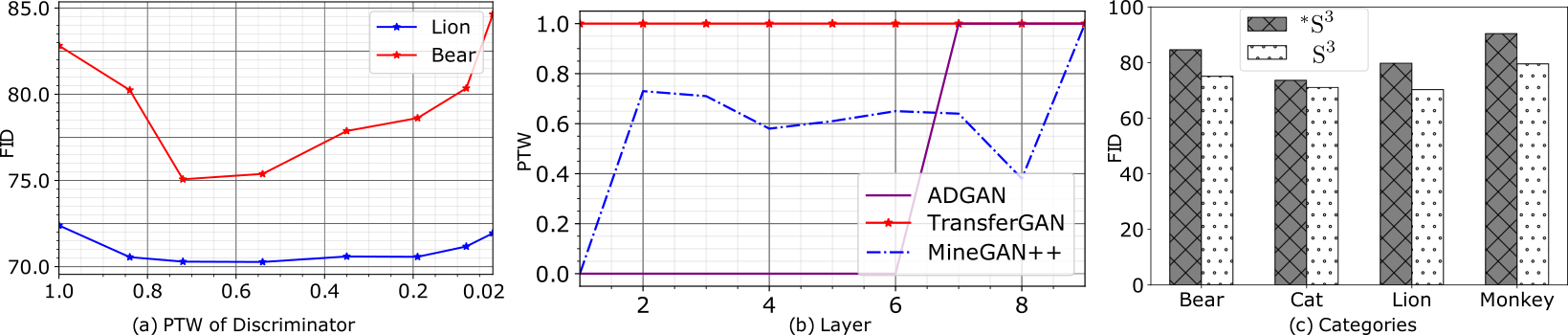}
    \caption{\small (a) FID as a function of the percentage of the trainable weight (PTW) when using the pretrained StyleGAN. (b) The percentage of trainable weights of each layer when using the pretrained StyleGAN. (c) Comparison between Sparse Subnetwork Selection  $\mathrm{S}^3$ and  *$\mathrm{S}^3$~(selecting weights instead of  filters). This shows the importance of the proposed filter selection.} 
    \label{fig:percentage_weight_on_layers}
\end{figure*}

\begin{figure*}[t]
    \centering
    \includegraphics[width=\textwidth]{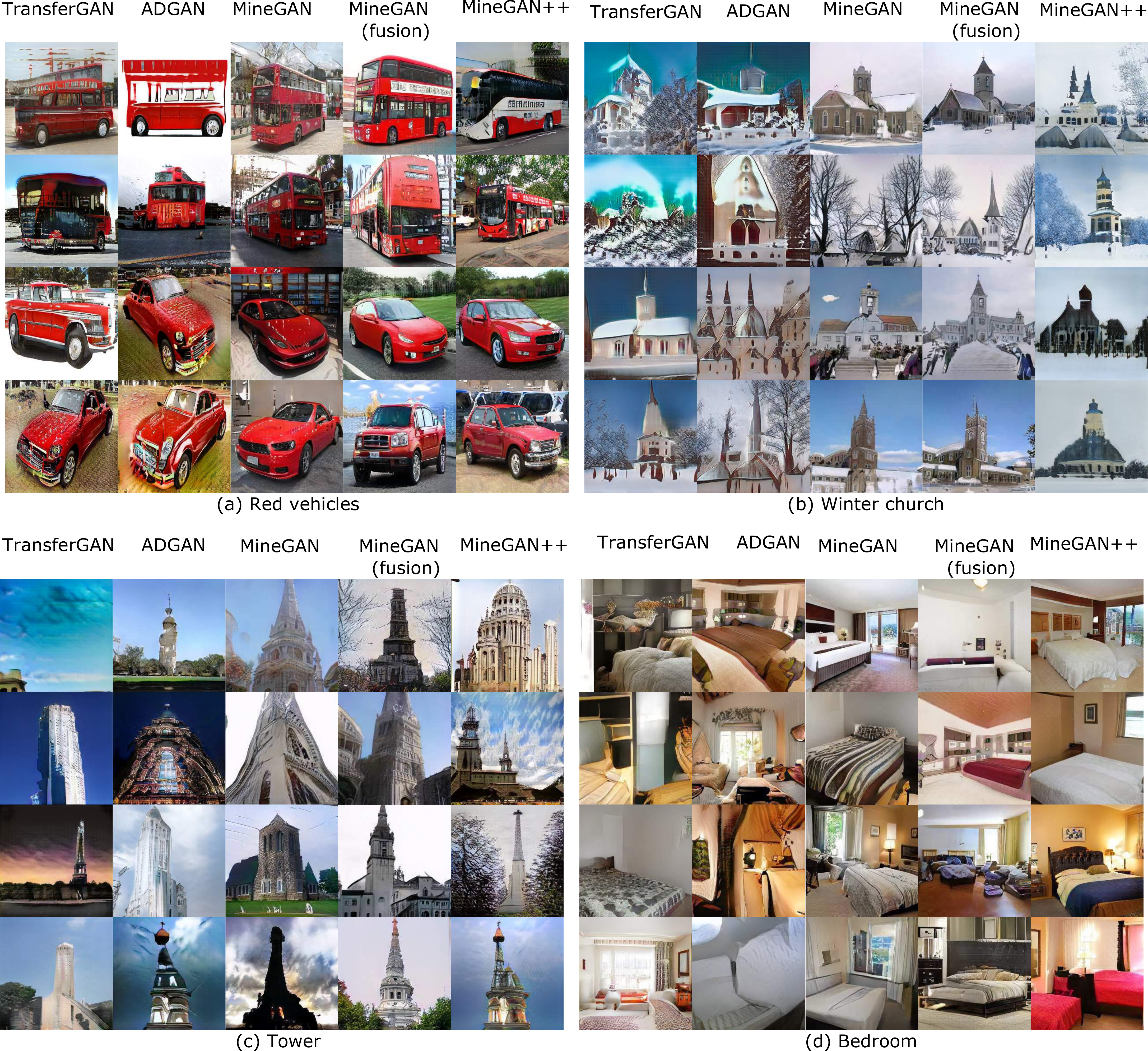}
    \caption{\small Results for transferring from multiple pretrained GANs. (a) $\{$car, bus$\}$ $\rightarrow $ Red vehicle, (b) $\{$Winter, Church$\}$ $\rightarrow $ Winter church, (c) $\{$Livingroom, Bridge, Church, Kitchen$\}$ $\rightarrow $ Tower   and (d) $\{$Livingroom, Bridge, Church, Kitchen$\}$ $\rightarrow $ Bedroom. All results are based on pretrained Progressive GANs.
    }
    \label{fig:target_vehicles}
    
\end{figure*}

\begin{table*}[t]
    \centering
        {\setlength{\tabcolsep}{20pt}\renewcommand{\arraystretch}{0.1}

          \resizebox{\textwidth}{!}{
            \centering
            \begin{tabular}{lcccc}
                \toprule
                  Method & $\rightarrow$ Red vehicle & $\rightarrow$ Winter church & $\rightarrow$ Tower & $\rightarrow$ Bedroom \\ 
                  \midrule
                  Scratch & 190 / 185 / 196 & 257 & 176 & 181\\
                  TransferGAN (bus) & 72.8 / 71.3 / 73.5 & - & - & - \\
                  ADGAN (bus) & 62.9 / 62.7 / 60.1 & - & - & -\\

                  TransferGAN (livingroom) & - & -& 78.9 & 65.4 \\
                  TransferGAN (church) & -& 85.4 & 73.8 & 71.5 \\
                  ADGAN (livingroom) & - & -& 65.9 & 52.1 \\
                  ADGAN (church) & -& - & 64.5 & 55.6 \\
                  MineGAN(multi)~\citep{wang2020minegan} & 61.2 / 59.4 / 61.5& 80.4 &62.4 & 54.7\\
                 MineGAN(multi) & 53.7 / 53.2 / 54.6& 75.4 &61.3 & 51.7\\

                  MineGAN(fusion) & 57.2 / 55.8 / 56.1& 69.3 &60.2 & 49.7\\
                  MineGAN++(fusion) & \textbf{49.9} / \textbf{48.5} / \textbf{49.6}& \textbf{62.7} &\textbf{55.3} & \textbf{44.2}\\

                  \midrule
                  \midrule
                  Estimated $\pi_i$ & & & \\
                  \midrule
                  Car & 0.31 / 0.46 / 0.63& - & - & - \\ 
                  Bus & 0.69 / 0.54 / 0.37& - & - & - \\
                  Livingroom & -& - & 0.08 & 0.46 \\
                  Kitchen & -& - & 0.07 & 0.40 \\
                  Bridge & -& - & 0.41 & 0.08 \\
                  Church & -& 0.48 & 0.44 & 0.06 \\
                  Winter & -& 0.52 & - & - \\
                \bottomrule  
            \end{tabular}
            }}
            \caption{\small Results for $\{$Car, Bus$\}$ $\rightarrow $ Red vehicle with three different target data distributions (ratios cars:buses are 0.3:0.7, 0.5:0.5 and 0.7:0.3) and $\{$Livingroom, Bridge, Church, Kitchen$\}$ $\rightarrow $ Tower/Bedroom. 
    (Top) FID scores between real and generated samples. (Bottom) Estimated probabilities $\pi_i$ for MineGAN(multi). 
    For the methods TransferGAN and ADGAN, we only show the result for that pretrained model that achieves the best score. MineGAN(multi)~\citep{wang2020minegan} refers to the implementation in Eqs.~\ref{eq:d_loss} and \ref{eq:g_loss},  and MineGAN(multi) to the improved version of Eqs.~\ref{eq:d_loss_permutate} and~\ref{eq:g_loss_permutate}. }  \label{table:lsun_fid_pro}
\end{table*}

We report the results in Table~\ref{tab:fid_animal}. Training the model from scratch obtains catastrophic scores on all target domains, indicating that using small datasets to train StyleGAN is impossible, since the discriminator suffers from overfitting and fails to provide meaningful information to update the generator. Benefiting from the prior information learned by the pretrained model, TransferGAN largely improves the performance.
Both FreezeD and ADGAN~\citep{zhao2020leveraging}  further improve the performance by fixing several layers of the model. In most cases, MineGAN gets better performance compared with the others (TransferGAN, FreezeD and ADGAN).
We also implemented Sparse Subnetwork Selection in TransferGAN (referred to as TransferGAN++), showing that the approach is general, does not require mining, and can benefit other approaches beyond MineGAN. Finally, \textit{MineGAN++} achieves the best score, again demonstrating that adaptively selecting the trainable filters outperforms simply manually fixing the low-level layers~\citep{mo2020freeze, zhao2020leveraging}.   {It should be noted that for two of the classes (\textit{Deer} and \textit{Dog}) MineGAN obtains inferior results, showing that the method is unstable for some target domains.}

Fig.~\ref{fig:stylegan_animal} show the synthesized image of the proposed method (MineGAN++) on target animal domains (\textit{cat}, \textit{lion} and \textit{wolf}). We note that our method successfully generates the target images.

Fig.~\ref{fig:percentage_weight_on_layers}(a) reports the effect of the trainable weight of the discriminator for StyleGAN on the target lion dataset. Notably, the model obtains the best score when using 54.8\% of all weights. Intuitively, one should leverage less parameters to adapt to the target domain.

In order to probe the difference of the discriminator devised by baselines and our method, we show the portion of the trainable weights at each layer of the discriminator of the StyleGAN obtained by Sparse Subnetwork Selection in Fig.~\ref{fig:percentage_weight_on_layers}(b). We observe that the three methods present different behavior. Specially, TransferGAN~\citep{wang2018transferring} updates all parameters to learn the target domain, thus is prone to suffer from ovefitting. ADGAN~\citep{zhao2020leveraging} only update the top layers. While the proposed method adaptively probes the weights on different layer. One interesting point is that the algorithm decides that the first layer (ResnetBlock layer) should be frozen, which presents the low-level image information (e.g., color, pattern etc..) of the feature extractor. Also, note that the algorithm chooses to totally update the last layer (the fully connection layer).

\begin{figure*}[t]
    \centering
    \includegraphics[width=\textwidth]{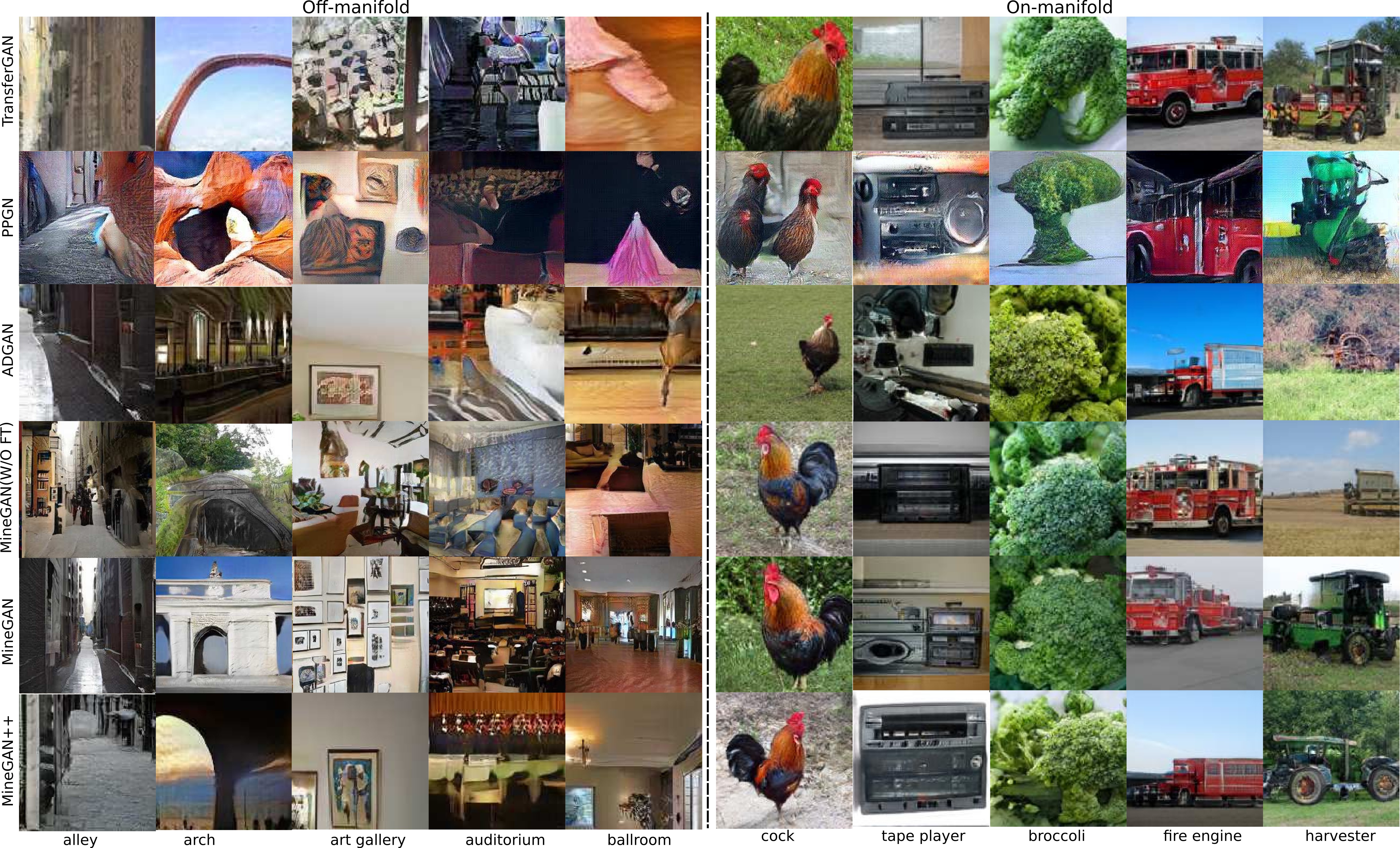}
    \caption{\small Results for conditional GAN. (Left) Far-domain (ImageNet$\rightarrow$Places365). (Right) Close-domain (ImageNet$\rightarrow$ImageNet).} 
    \label{fig:conditional_imagenet_places}
\end{figure*}

\subsubsection{Multiple pretrained models}
\label{sec:exp_mult_gans}
We now evaluate the case where there is more than one pretrained model to mine from. We leverage two groups of pretrained Progressive GANs: (a) the \textit{car} and \textit{bus} model, (b)  the \textit{winter} and \textit{church} model. We train the models for the car, bus, and church categories by using corresponding images from the LSUN dataset~\citep{yu2015lsun}, while we train the \textit{winter} model with a collection of 10,000 images of winter scenes, which is got by Google. 

We provide results for the two methods  that we propose to transfer knowledge from multiple GANs. We use \emph{MineGAN(multi)} to denote the method that mines from multiple GANs (see Section~\ref{sec:multipleGANs}). We also show the results of the fused model, which combines the separate pretrained GANs into a single conditional GAN (see Section~\ref{sec:fusion}). We then apply mining to the fused model. These results are indicated with the name \emph{MineGAN(fusion)}. When we compare to methods that can only be based on a single pretrained GAN (TransferGAN and ADGAN) we run with all pretrained models, and report only the best results.

\minisection{Red vehicle dataset} For both the \textit{car} and \textit{bus} models, 
 these pretrained networks generate cars and buses of a variety of different colors. We collect a target dataset of 200 images of \emph{red vehicle} (with resolution $256 \times 256$), which contains both red cars and red buses.

\minisection{Winter church dataset} For both the \textit{winter} and \textit{church} models, we aim to adapt them to the target data: \textit{Winter church}. We collect 200 target images (of $256 \times 256$ resolution). 
In this setting, each of the target images has an overlap with both pretrained domains of \textit{winter}  and \textit{church} images. Differently to the \textit{red vehicle} images, here for each target image knowledge from both pretrained models is helpful. This setting is expected to be beneficial to the mining from conditional GANs explained in Sec.~\ref{sec:fusion}.

Table~\ref{table:lsun_fid_pro}~(top rows) provides the quantitative results.  On the \emph{Red vehicle} dataset, we consider three target sets with different car-bus ratios (0.3:0.7, 0.5:0.5, and 0.7:0.3) which allows us to evaluate the estimated probabilities $\pi_i$ of the selector. The results show that both MineGAN(multi) and MineGAN(fusion) outperfrom TransferGAN with a significantly lower FID score. We also show that our improved optimization (see Eqs.~\ref{eq:d_loss_permutate} and~\ref{eq:g_loss_permutate}) lowers the FID from 59.4
of the original MineGAN(multi)~\citep{wang2020minegan} to 53.2 (for the ratio cars:buses is 0.5:0.5). Furthermore, the probability distribution predicted by the selector, reported in Table~\ref{table:lsun_fid_pro}~(bottom rows), matches the class priors of the target data. Finally, combining both Sparse Subnetwork Selection and GAN fusion, we obtain the best FID score of 48.5.

The results on \emph{Winter church} confirm the results we obtain on \emph{red vehicle} and again our MineGAN approaches improve significantly over the TransferGAN. The main difference is that on this dataset the results with MineGAN(fusion) are better than those based on MineGAN(multi); the FID score drops significantly from 80.4 to 75.4. As discussed, this is expected, since for the correct generation of a single Winter church image, knowledge from both pretrained GANS is beneficial. Again, the best results are obtained with MineGAN++(fusion), leading to a further drop of more than 6 FID points.

Fig.~\ref{fig:target_vehicles}~(a,b) shows the synthesized images. As expected, the limited amount of data makes training from scratch result in overfitting.   
Both TransferGAN~\citep{wang2018transferring} and ADGAN~\citep{zhao2020leveraging} produce only high-quality output samples for one of the two classes (the class that coincides with the pretrained model) as they cannot extract knowledge from both pretrained GANs. On the other hand, MineGAN generates high-quality images by successfully transferring the knowledge from both source domains simultaneously.  MineGAN (fusion) further improves the performance, especially for the winter-church image.  {Finally, MineGAN++ has the best visual quality.}

\begin{table*}[t]
    \centering
    {\setlength{\tabcolsep}{20pt}\renewcommand{\arraystretch}{1}
    \begin{subtable}{1\linewidth}
      \centering
          \resizebox{1\columnwidth}{!}{
            \begin{tabular}{lccccc}
            \toprule
                            \multirow{2}{*}{Method}  & \multicolumn{2}{c}{ Far-domain}  & \multicolumn{2}{c}{ Close-domain}& \multirow{2}{*}{Generation time (ms)}\\
                            & Label & FID/KMMD & Label & FID/KMMD & \\ 
                    \midrule
                 Scratch & No &190 / 0.96&No&187 / 0.93&  {\textbf{5.1}} \\
                 TransferGAN & No &89.2 / 0.53&No&58.4 / 0.39&  {\textbf{5.1}} \\
                 DGN-AM & Yes &214 / 0.98&Yes&180 / 0.95&3020\\ 
                 PPGN& Yes & 139 / 0.56&Yes&127 / 0.47&3830\\ 
                 ADGAN & No &85.4 / 0.51&No&54.2 / 0.28&  {\textbf{5.1}} \\
                 MineGAN (w/o FT)& No& 82.3 / 0.47 &No&61.8 / 0.32&5.2 \\ 
                 MineGAN& No& 78.4 / 0.41 &No&52.3 / 0.25&5.2 \\ 
                 MineGAN++& No&   {\textbf{74.3}} /   {\textbf{0.40}} &No&  {\textbf{50.4}} /   {\textbf{0.24}}&5.2 \\ 
                \bottomrule
            \end{tabular}
            }

    \end{subtable} }
    \caption{\small  Distance between real data and generated samples as measured by FID score and KMMD value. The Far-domain results correspond to ImageNet $\rightarrow $ Places365, and the Close-domain results correspond to ImageNet $\rightarrow $ ImageNet. We also indicate whether the method requires the target label. Finally, we show the inference time for the various methods in milliseconds.}
    \label{table:appendix_target_imagenet_places365_fid}
\end{table*}

\minisection{Bedroom and Tower datasets} To demonstrate the scalability of MineGAN with multiple pretrained models, we conduct experiments using four different generators, each trained on a different LSUN category including \textit{livingroom}, \textit{kitchen}, \textit{church}, and \textit{bridge}.
We consider two different Far-domain target datasets, one with \textit{bedroom} images and one with \textit{tower} images, both containing 200 images. 
Table~\ref{table:lsun_fid_pro}~(top) shows that our method obtains significantly better FID scores even when we choose the most relevant pretrained GAN to initialize training for TransferGAN. The conclusions are similar to previous experiments: both MineGAN(multi) and MineGAN(fusion) obtain similar improved results, whereas the best results are obtained with the Sparse Subnetwork Selection. In general, on all datasets we see that we improve the FID score by almost 20 points when compared with the naive finetuning approach (TransferGAN).

Table~\ref{table:lsun_fid_pro}~(bottom) shows that the miner identifies the relevant pretrained models, e.g. transferring knowledge from \textit{bridge} and \textit{church} for the target domain \textit{tower}. Finally, Fig.~\ref{fig:target_vehicles}~(c,d) provides visual examples.

\minisection{Ablation initialization strategy} We provide additional results for the experiment ($\{$Bus, Car$\}$) $\rightarrow $ Red vehicle and for the experiment  $\{$Bedroom, Bridge, Church, Kitchen$\}$ $\rightarrow $ Tower/Bedroom.
When applying MineGAN to multiple pretrained GANs, we use one of the domains to initialize the weights of the critic. We used 
\textit{church} to initialize the critic in case of the target set \textit{tower}, and \textit{kitchen} to initialize the critic for the target set \textit{bedroom}. We found this choice to be of little influence on the final results. When using \textit{kitchen} to initialize the critic for target set \textit{tower}, the FID score changes from 61.3 to 62.9. 
When using \textit{church} to initialize the critic for target set \textit{bedroom} results change from 51.7 to 52.2.

\subsection{Knowledge transfer from conditional GANs}
Here we transfer knowledge from a pretrained \emph{conditional GAN} (see Section~\ref{sec:condition}).
We use BigGAN~\citep{brock2018large}, which is trained using ImageNet~\citep{russakovsky2015imagenet}, and evaluate on two target datasets: Close-domain (ImageNet: \textit{cock}, \textit{tape player}, \textit{broccoli},  \textit{fire engine},  \textit{harvester}) and Far-domain (Places365~\citep{zhou2014object}:  \textit{alley},  \textit{arch},  \textit{art gallery},  \textit{auditorium}, \textit{ballroom}). 
We use 500 images per category. We compare our method with training from scratch, TransferGAN~\citep{wang2018transferring}, ADGAN~\citep{zhao2020leveraging}, and two iterative methods: DGN-AM~\citep{nguyen2016synthesizing} and PPGN~\citep{nguyen2017plug} \footnote{We were unable to obtain satisfactory results with BSA~\citep{noguchi2019image} in this setting (images suffered from  blur artifacts) and have excluded it here.}.

It should be noted that both DGN-AM~\citep{nguyen2016synthesizing} and  PPGN~\citep{nguyen2017plug} are based on a less complex GAN (equivalent to DCGAN~\citep{radford2015unsupervised}).
Therefore, we expect these methods to exhibit results of inferior quality, and so the comparison here should be interpreted in the context of GAN quality progress. However, we would like to stress that they do not aim to transfer knowledge to new domains. 
They can only generate samples of a particular class of a pretrained classifier network, and they have no explicit loss ensuring that the generated images follow a target distribution. 

Fig.~\ref{fig:conditional_imagenet_places} shows qualitative results for the different methods.  
As in the unconditional case, both MineGAN++ and MineGAN  produce  very realistic results, even for the challenging Far-domain case. 
Table~\ref{table:appendix_target_imagenet_places365_fid} presents quantitative results in terms of FID and KMMD.
We also indicate which methods use the label of the target domain class.
MineGAN++ obtains the best scores for both metrics, despite not using target label information. ADGAN achieves a similar score to MineGAN for \textit{Close-domain}, but inferior performance for \textit{Far-domain}.
PPGN performs significantly worse than our method.
TransferGAN has a large performance drop for \textit{Far-domain} case, for which it cannot use the target label as it is not in the pretrained GAN (see~\citep{wang2018transferring} for details). 

Another important point regarding DGN-AM and PPGN is that each image generation during inference is an iterative process of successive backpropagation updates until convergence, whereas our method is feedforward.
Therefore, we include in Table~\ref{table:appendix_target_imagenet_places365_fid} the inference running time, using the default 200 iterations for DGN-AM and PPGN. 
All timings have been computed with a CPU Intel Xeon E5-1620 v3 @ 3.50GHz and GPU NVIDIA RTX 2080 Ti.
We can clearly observe that the feedforward methods (TransferGAN, ADGAN and MineGAN) are three orders of magnitude faster despite being applied on a more complex GAN~\citep{brock2018large}.

\section{Conclusions}\label{sec:conclusions}
This paper aims to leverage the rich knowledge available in generative adversarial networks trained on large datasets, so they can be effectively and efficiently adapted to target domains with a limited amount of training images.
This allows to also exploit GANs on domains for which few data are available
and it significantly reduces data collection costs. 
For this reason, we propose mining, a new knowledge transfer method that effectively transfers knowledge from single or multiple GAN domains to small target domains. 
Mining involves an additional small network that identifies those regions on the learned GAN manifold that are closer to a given target domain. Mining leads to more  efficient finetuning, especially with few target domain images. Moreover, to further prevent overfitting, we proposed Sparse Subnetwork Selection, a technique that identifies a subset of parameters in the discriminator that are relevant for the target class. Restricting the optimization to these neurons further reduces the chance of overfitting, and is shown to result in improved image quality. Experiments with various GAN architectures (BigGAN, StyleGAN, and SNGAN) on multiple datasets demonstrated that MineGAN obtains state-of-the-art results. Finally, we are the first to demonstrate results for knowledge transfer from multiple pretrained GANs.

\section{Data Availability Statements}
All the data for used during the study are openly available from the corresponding author upon reasonable request. The training data are got access on related works~\citep{karras2019style,choi2020stargan,wang2020minegan,zhu2017unpaired,deng2009imagenet,yu15lsun}.

\begin{acknowledgements}
We acknowledge the support from Huawei Kirin Solution, and the Grant PID2019-104174GB-I00 funded by MCIN/AEI/ 10.13039/501100011033 and Grant PID2021-128178OB-I00 funded by MCIN/AEI/ 10.13039/501100011033 and by ERDF A way of making Europe, Ramón y Cajal fellowship Grant RYC2019-027020-I funded by MCIN/AEI/ 10.13039/501100011033 and by ERDF A way of making Europe, and the CERCA Programme of Generalitat de Catalunya.  We acknowledge the project supported by Youth Foundation (62202243).

\end{acknowledgements}

\bibliographystyle{spbasic}      %
\bibliography{shortstrings,refs}

\appendix

  {
\section{Architecture and training details}\label{appdendx_sec:architecture_and_training}
\textbf{MNIST dataset.} Our model contains a \textit{miner}, a \textit{generator} and a \textit{discriminator}. For both unconditional and conditional GANs, we use the same framework~\citep{gulrajani2017improved} to design the generator and discriminator. The miner is composed of two fully connected layers with the same dimensionality as the latent space $|z|$. The visual results are computed with $|z|=16$; we found that the quantitative results improved for larger $|z|$ and choose $|z|=128$.

We randomly initialize the weights of each miner following a Gaussian distribution centered at 0 with 0.01 standard deviation, and optimize the model using Adam~\citep{kingma2014adam} with batch size of 64. The learning rate of our model is 0.0004,  with exponential decay rates of $\left ( \beta_{1},   \beta_{2} \right ) = \left ( 0.5, 0.999 \right )$.

\textbf{CelebA Women,  FFHQ Children and LSUN (Tower and Bedroom) datasets.} We design the generator and discriminator based on Progressive GANs~\citep{karras2017progressive}. Both networks use a multi-scale technique to generate high-resolution images. Note we use a \emph{simple} miner for tasks with dense and narrow source domains. The miner comprises out of four fully connected layers (8-64-128-256-512), each of which is followed by a \textit{ReLU} activation and \textit{pixel normalization}~\citep{karras2017progressive} except for last layer.  We use a Gaussian distribution centered at 0 with 0.01 standard deviation to initialize the miner, and optimize the model using Adam~\citep{kingma2014adam} with batch size of 4. The learning rate of our model is 0.0015, with exponential decay rates of $\left ( \beta_{1},   \beta_{2} \right ) = \left ( 0, 0.99 \right )$. 

\textbf{Animal dataset.} We use StyleGAN~\citep{karras2019style} as our pretrained model. In StyleGAN, the generator contains a mapping network and a synthesis network, we thus devise two minor networks.   The first one, which explores the latent space of the mapping network, is composed of two fully connected layers (512-512), in which the first one is followed by a \textit{RELU}. The last miner, which is corresponding to the synthesis network, consists of five fully connections (16-16-64-256-1024), each of which is followed by a \textit{RELU} except for the last one.  We use Adam~\citep{kingma2014adam} with a batch size of 2, following a hyper parameter learning rate of 0.00001 and  exponential decay rate of  $\left ( \beta_{1},   \beta_{2} \right ) = \left ( 0, 0.99 \right )$.  

\textbf{FFHQ Face and Anime Face datasets.} We use the same network as~\citep{miyato2018spectral}, namely SNGAN. The miner consists of three fully connected layers (8-32-64-128).  We randomly initialize the weights following a Gaussian distribution centered at 0 with 0.01 standard deviation.
 For this additional set of experiments, we use Adam~\citep{kingma2014adam} with a batch size of 8, following a hyper parameter learning rate of 0.0002 and  exponential decay rate of  $\left ( \beta_{1},   \beta_{2} \right ) = \left ( 0, 0.9 \right )$. 

\textbf{Imagenet and Places365 datasets.}  We use the pretrained BigGAN~\citep{brock2018large}. We ignore the projection loss in the discriminator, since we do not have access to the label of the target data.   We employ a more \emph{powerful} miner in order to allocate more capacity to discover the regions related to target domain. The miner consists of two sub-networks:  miner $M^z$ and  miner $M^c$. Both $M^z$ and  $M^c$ are composed of four fully connected layers of sizes (128, 128)-(128, 128)-(128, 128)-(128, 128)-(128, 120) and (128, 128)-(128, 128)-(128, 128)-(128, 128)-(128, 128), respectively. 
We use Adam~\citep{kingma2014adam} with a batch size of 256, and learning rates of 0.0001 for the miner and the generator and 0.0004 for the discriminator. 
The exponential decay rates are $\left ( \beta_{1},   \beta_{2} \right ) = \left ( 0, 0.999 \right )$. We randomly initialize the weights following a Gaussian distribution centered at 0 with 0.01 standard deviation.

The input of BigGAN is a random vector and a class label that is mapped to an embedding space. 
We therefore have two miner networks, the original one that maps to the input latent space ($M^z$) and a new one that maps to the latent class embedding ($M^c$). Note that since we have no class label, the miner ($M^c$) needs to learn what distribution over the embeddings best represents the target data.

\section{FID computation}\label{appdendx_sec:demension_fid}
We compute the FID for the pretrained Stylegan and Styleganv2 based on different amounts of test images.   Note here, both StyleGAN and styleGANv2 are trained on HHFQ,  and tested on HHFQ. When computing FID, the number of both the test images and the generated images is the same. We repeat this 10 times for each test dataset, and report the mean value.  As shown in Figure~\ref{fig:fid_demension}(top), even though with the varying test images both methods have different values,  the trend is consistent that StyleGANv2 always outperforms StyleGAN even when only using a 100 images. In conclusion, given the same number of images when comparing methods, the FID still provides a good measure to assess the relative quality.

We also compare our method with TransferGAN on AFHQ with 100 training images by using the pretrained styleganv2. As shown in Figure~\ref{fig:fid_demension}(bottom), we are able to draw a similar conclusion: our method still outperforms the other when using limited data to evaluate models.

\begin{figure}[t]
    \centering
    \includegraphics[width=0.9\columnwidth]{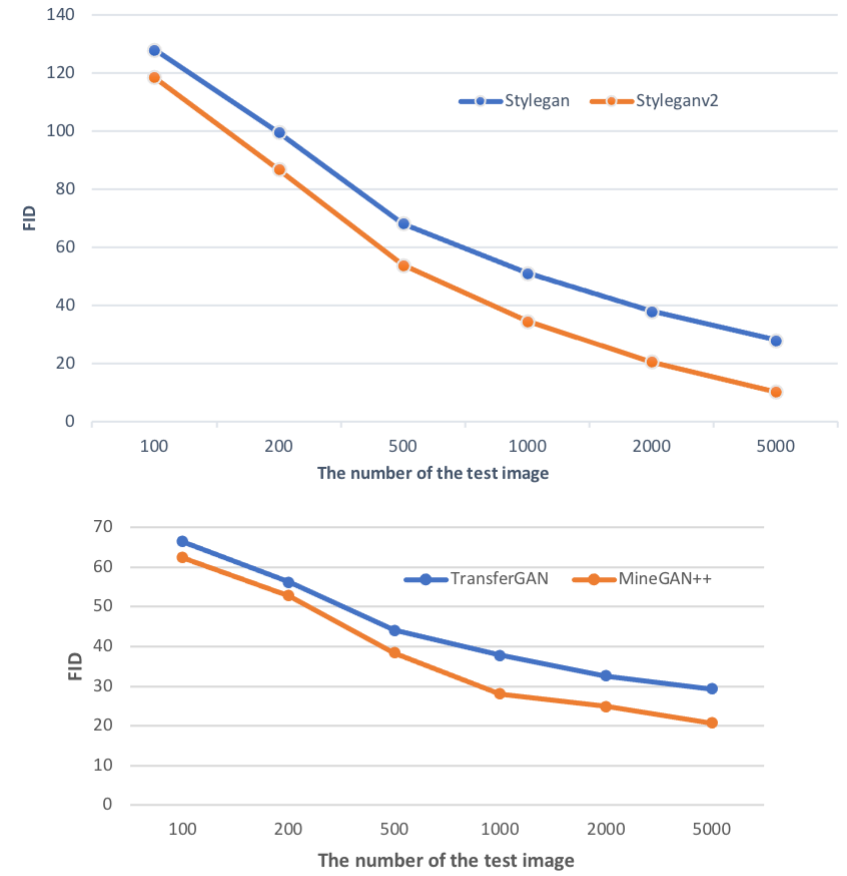}
    \caption{\small   {The FID as a function of the number of the image used for its computation. (Top) The FID for  the pretrained Stylegan and Styleganv2; (bottom) The FID for the baseline TransferGAN and MineGAN++. Results show that the ranking based on FID differences remains consistent when varying the number of images used to measure it.}}
    \label{fig:fid_demension}
\end{figure}

\begin{figure*}[t]
    \centering
    \includegraphics[width=\textwidth]{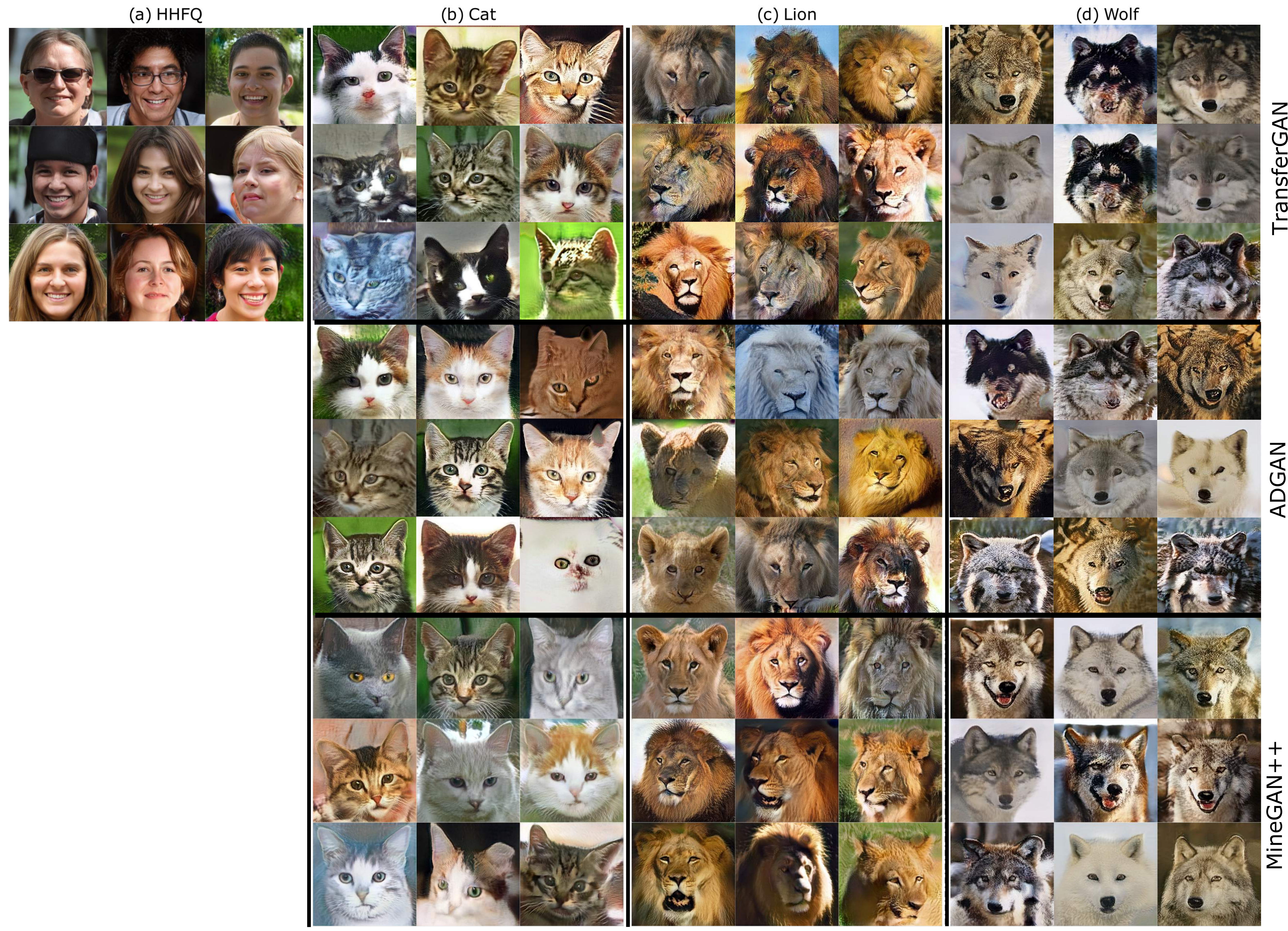}
    \caption{\small   {Comparing results between MineGAN++ and the baselines.}}
    \label{fig:append_animals_more}
\end{figure*}

\section{Ablation with different pretrained models}
When applying MineGAN to multiple pretrained GANs, we use one of the domains to initialize the weights of the critic. In \textbf{Table 5} we used 
\textit{Church} to initialize the critic in case of the target set \textit{Tower}, and \textit{Kitchen} to initialize the critic for the target set \textit{Bedroom}. We found this choice to be of little influence on the final results. When using \textit{Kitchen} to initialize the critic for target set \textit{Tower} results change from 62.4 to 61.7. When using \textit{Church} to initialize the critic for target set \textit{Bedroom} results change from 54.7 to 54.3.

\section{Additional comparison results}

we have  performed transfer learning from human face to Pigeon. We collect 200 pigeion images with $256 \times 256$. We finetune StyleGAN. As shown on Fig.~\ref{fig:append_pigeon},  although MineGAN(w/o FT) obtains unsatisfying results,  we could improve our method with further finetunning. We evaluate the baselines and our method, and report FID: (TGAN, ADGAN, MineGAB(w/o FT), MineGAN, MineGAN++): (181,178, 197, 173, 169), which again verifies the proposed method has better performance than the baselines.

\begin{figure*}[t]
    \centering
    \includegraphics[width=\textwidth]{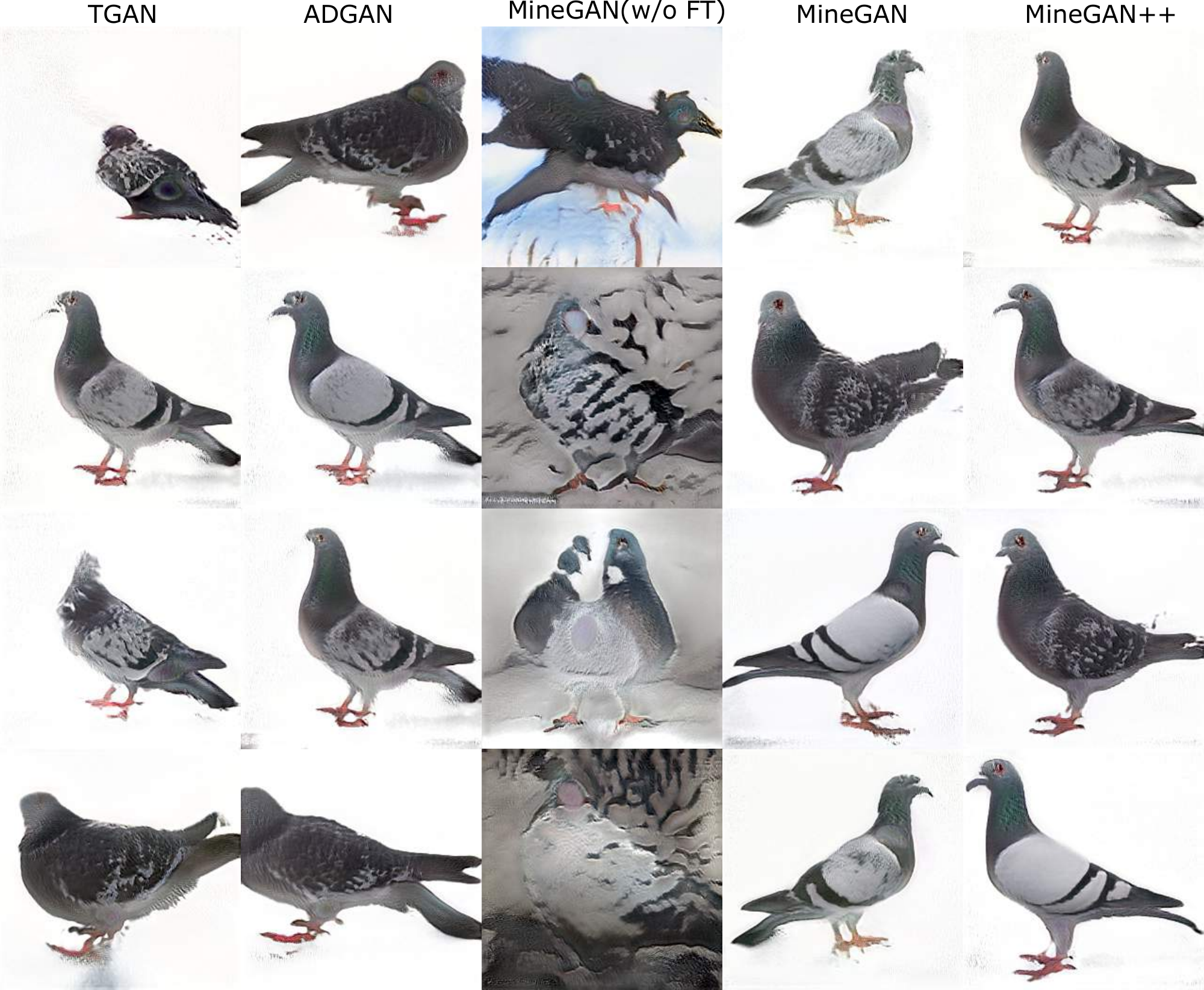}
    \caption{\small   {Comparison results between MineGAN++ and the baselines.}}
    \label{fig:append_pigeon}
\end{figure*}

\section{Ablation of sparse subnetwork selection}
  {We use the gradient norm (see Eq.~\ref{eq:compar}) for the selection of the sparse subnetwork. However, other criteria could also be considered. Here,
we also consider method Kernel weights~\citep{han2015learning} and BN statistics~\citep{liu2017learning} for the selection of the sparse subnetwork.
We have  added an experiment  on animal faces: Bear, Monkey and Lion. As reported in Tab.~\ref{tab:ablation_selection_methods}, the other criteria can also improve the model performance in most cases. However, our proposed method, based on the gradient norm, obtains the best FID score, which indicates that using gradient norm is better than naive magnitude of kernel weights or BN statistics. The reason is that using the gradient,  computed by the target data, is a more accurate indicator of the important filters. }

\begin{table}[t]
\centering
	\def\arraystretch{0.5}
    \begin{adjustbox}{max width=\textwidth}
           \resizebox{1\linewidth}{!}{
            \begin{tabular}{lccc}
    \toprule
    \multirow{1}{*}{} & \multicolumn{1}{c}{Bear} & \multicolumn{1}{c}{Monkey}& \multicolumn{1}{c}{Lion} \\
     \midrule 
    Kernel weights~\citep{han2015learning}
 & 76.24 &81.43&71.52 \\
    BN statistics~\citep{liu2017learning} & 75.31&80.27&71.57\\
   MineGAN & 76.53 & 82.10&72.39
 \\ 
    MineGAN++ & \textbf{75.07}  &\textbf{79.53}&\textbf{70.27}
 \\ 
                \bottomrule
            \end{tabular}
            }
    \end{adjustbox}
    \caption{\small   {Ablation study with different  criterions to evaluate the importance of each filter.}} 
	\label{tab:ablation_selection_methods}
\end{table}

}

\end{document}